\pdfoutput=1

\documentclass[11pt]{article}

\usepackage[final]{acl}

\usepackage{times}
\usepackage{latexsym}
\usepackage{lipsum}
\usepackage{amsmath}
\usepackage{amssymb}

\usepackage[T1]{fontenc}

\usepackage[utf8]{inputenc}

\usepackage{microtype}

\usepackage{inconsolata}

\usepackage{graphicx}

\usepackage{adjustbox}      
\usepackage{booktabs}       
\usepackage{makecell}       
\usepackage{multirow}       
\usepackage{caption}        
\usepackage{amsmath}        
\usepackage{amssymb}        
\usepackage{colortbl}
\usepackage{xspace}
\usepackage{xcolor}
\usepackage{mathrsfs}
\usepackage{comment}
\usepackage{algorithm}
\usepackage{algorithmic}
\usepackage{soul}

\usepackage{subcaption}
\def \ie{\textit{i.e.}, }
\def \eg{\textit{e.g.}, }

\definecolor{figureyellow}{RGB}{237,182,7}
\definecolor{figurered}{RGB}{240,84,136}
\definecolor{figureblue}{RGB}{67,145,239}
\definecolor{figuregreen}{RGB}{74,200,68}

\title{\textit{Can LLMs Deceive CLIP?} Benchmarking Adversarial Compositionality of Pre-trained Multimodal Representation via Text Updates}

\author{
\quad \textbf{Jaewoo Ahn}$^{*}$
\quad \textbf{Heeseung Yun}$^{*}$
\quad \textbf{Dayoon Ko}
\quad \textbf{Gunhee Kim}\\
Seoul National University \\
\texttt{\small \{jaewoo.ahn, heeseung.yun, dayoon.ko\}@vision.snu.ac.kr, \small gunhee@snu.ac.kr} \\
\small{\url{https://vision.snu.ac.kr/projects/mac}} \\
}

\newcommand{\correspondingfootnote}{
    \let\oldthefootnote=\thefootnote
    \renewcommand{\thefootnote}{}
    \footnotemark
    \footnotetext{$^{*}$Equal Contribution}
    \let\thefootnote=\oldthefootnote
}

\begin{document}
\maketitle
\begin{abstract}
While pre-trained multimodal representations (\eg CLIP) have shown impressive capabilities, they exhibit significant compositional vulnerabilities leading to counterintuitive judgments.
We introduce Multimodal Adversarial Compositionality (MAC), a benchmark that leverages large language models (LLMs) to generate deceptive text samples to exploit these vulnerabilities across different modalities and evaluates them through both sample-wise attack success rate and group-wise entropy-based diversity.
To improve zero-shot methods, we propose a self-training approach that leverages rejection-sampling fine-tuning with diversity-promoting filtering, which enhances both attack success rate and sample diversity.
Using smaller language models like Llama-3.1-8B, our approach demonstrates superior performance in revealing compositional vulnerabilities across various multimodal representations, including images, videos, and audios.
\end{abstract}

\correspondingfootnote

\section{Introduction}
\label{sec:introduction}

Recent advances in multimodal systems have demonstrated remarkable capabilities in generating multimodal content from multimodal inputs.
At the core of these developments lies pre-trained multimodal representations, 
which can encode rich information from different modalities. 
Such representations, notably illustrated by Contrastive Image-Language Pre-Training (CLIP)~\cite{radford2021clip}, has become an indispensable component in modeling complex contextual understanding in crossmodal settings,
finding widespread applications across retrieval~\cite{luo2022clip4clip,ahn2023mpchat}, generation~\cite{ramesh2022unCLIP}, and reward modeling~\cite{yu2023esper,rocamonde2024vlmrm}. 
Moreover, 
its usage has become commonplace across various 
modalities beyond image-language pairs.

\begin{figure}[t]
    \centering
    \includegraphics[width=0.49\textwidth]{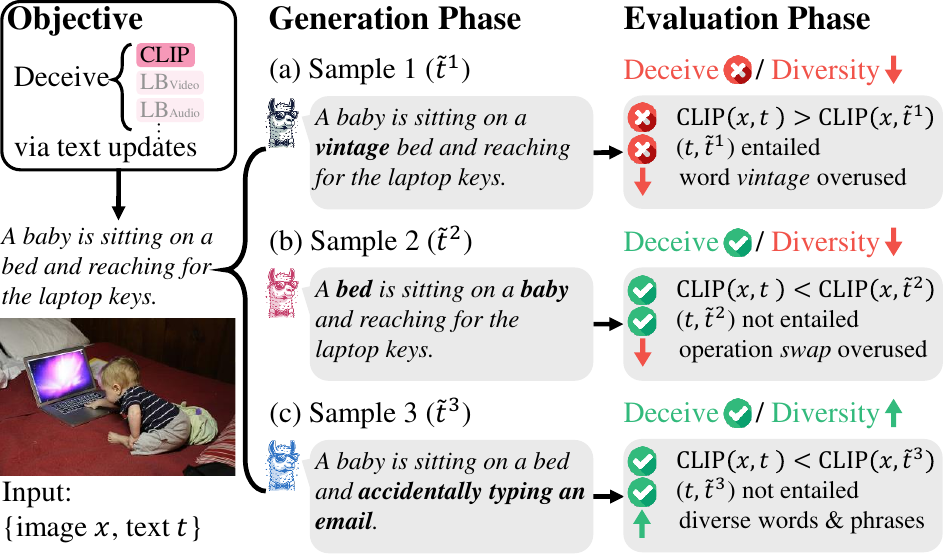}
    \caption{
Key idea of Multimodal Adversarial Compositionality (MAC).
MAC benchmarks compositional vulnerabilities of a pre-trained multimodal representation (\eg CLIP, LanguageBind) with a comprehensive set of criteria.
$\text{CLIP}(\cdot,\cdot)$ denotes the cosine similarity between image and text embeddings from CLIP.
}
    \label{fig:keyidea}
\end{figure}

Contrary to their prevalence in a wide range of downstream applications, pre-trained multimodal representations are known to be considerably brittle.
This brittleness can be intuitively exemplified by compounding text elements.
As illustrated in Fig.~\ref{fig:keyidea}-(b), with an image of a baby sitting, these systems may assign a high similarity score to an erroneous description like ``a \textit{bed} is sitting on a \textit{baby}'' than the correct description.
Such counterintuitive judgments occur surprisingly often, implying a critical issue where the vulnerabilities in the embeddings are inherited by the models that utilize them.
Consequently, there have been active efforts to identify these weaknesses through negative samples constructed from the perspective of visual compositional reasoning (\ie structured relationship between words and their corresponding visual elements), 
such as negation, event swapping, and attribute replacement~\cite{thrush2022winoground,ma2023crepe}.
However, developing a comprehensive understanding of diverse \textit{compositional vulnerabilities}, without assuming specific scenarios, remains an open challenge.

In this work, we introduce the challenge of large language models (LLMs) deceiving CLIP, 
\textit{i.e.}, exploiting weaknesses in how pre-trained multimodal representations encode relationships between objects and attributes in multimodal contents (\textit{e.g.}, image).
To this end, we propose to benchmark the \textbf{M}ultimodal \textbf{A}dversarial \textbf{C}ompositionality (MAC) of a target representation.
Given multimodal data pairs (\textit{e.g.}, image-caption), LLMs generate deceptive captions by slightly modifying ground-truth captions in a way that misaligns or contradicts the original content. 
We then rigorously evaluate whether the target representation mistakenly prefers these generated captions over the original ones. 
Unlike previous studies that address compositionality within specific modalities~\cite{thrush2022winoground,bansal2024videocon,ghosh2024compa}, our work highlights a key distinction in deceiving a target representation in a modality-agnostic manner (\textit{e.g.}, image, video, audio).

For evaluation, given a set of captions generated by LLMs for deceiving, we propose a testbed that assesses their effectiveness through \textit{sample-wise} and \textit{group-wise} evaluation.
We first evaluate whether each generated sample successfully executes an attack (\textit{sample-wise}).
This success requires meeting multifaceted conditions: the generated deceptive sample should (i) maintain high crossmodal similarity with the original multimodal input, (ii) contain non-entailing content while (iii) maintaining lexical similarity to the original text, and (iv) adhere to prescribed instructions without relying on shortcuts.
Furthermore, if they are predictable or monotonous, they become easily defensible and fail to unravel \textit{diverse} compositional vulnerabilities.
Therefore, we design entropy-based metrics to measure the diversity of composition elements used in deception across the set of generated samples (\textit{group-wise}).

In addition, we leverage the self-training of LLMs \cite{huang2023selfimprove}, particularly rejection sampling fine-tuning~\cite{touvron2023llama2} for the first time, where generated samples are used for additional training to promote deceptive response generation.
Existing zero-shot sample generation for compositionality and na\"ive self-training methods 
often fail to elicit diverse compositions using a limited set of elements.
To address this limitation, we propose a \textit{diversity-promoting} self-training approach by thorough sampling among sample candidates.
Even with smaller LLMs centered around Llama-3.1-8B \cite{dubey2024llama3}, our simple yet effective framework can substantially improve both attack success rates and diversity.
We achieve superior deception performance compared to prior work across various representations for multiple modalities, including image, video, and audio.
In particular, our method outperforms existing approaches~\cite{yarom2023SeeTrue,momeni2023verbsinaction,ghosh2024compa}, when evaluated on COCO~\cite{lin2014coco}, MSRVTT~\cite{xu2016msrvtt}, and AudioCaps~\cite{kim2019audiocaps}, successfully deceiving target models, notably CLIP~\cite{radford2021clip} and LanguageBind~\cite{zhu2023languagebind}.


\begin{table*}[t!]
\begin{center}
\begin{adjustbox}{width=\linewidth}
    \begin{tabular}{llllcccc}
    \toprule
    \multirow{2}{*}{Method} & Modality & \multirow{2}{*}{Generation} & Text Update & \multicolumn{4}{c}{Compositionality Criteria} \\
    & \small\makecell{(\textbf{I}mage, \textbf{V}ideo, \textbf{A}udio)} & & \small\makecell{(\textbf{R}eplace, \textbf{S}wap, \textbf{A}dd)} & Crossmodal  & Unimodal & Lexical & Diversity\\
    \midrule
    FOIL~\cite{shekhar2017foil}             & I & Rule-based & Specific (R) & E, F & F & F & - \\
    Winoground~\cite{thrush2022winoground}  & I & Human-annotated  & Specific (S) & E, F & F & F & -\\
    VL-CheckList~\cite{zhao2022vlchecklist} & I & Rule-based & Specific (R) & E, F & F & F & - \\
    RoCOCO~\cite{park2024rococo}            & I & Rule-based & Specific (R) & E, F & F & F & - \\
    ARO~\cite{yuksekgonul2022aro}           & I & Rule-based & Specific (S) & E, F & F & F & - \\ 
    SVLC~\cite{doveh2023SVLC}               & I & Rule-based & Specific (R) & E, F & F & F & - \\
    CREPE~\cite{ma2023crepe}                & I & Rule + LLM & Specific (R, S, A) & E, F & F & F & - \\ 
    SugarCrepe~\cite{hsieh2023sugarcrepe}   & I & LLM (ChatGPT) & Specific (R, S, A) & E, F & F & F & - \\ 
    SeeTrue~\cite{yarom2023SeeTrue}         & I & LLM (PaLM) & General & E, F & F & - & - \\ 
    LLaVA-Score~\cite{li2024llavascore}     & I & LLM (GPT-4) & Specific (R, S) & E, F & F & F & - \\ 
    FSC-CLIP~\cite{oh2024FSCCLIP}           & I & Rule-based & Specific (R, S) & E, F & F & F & - \\ 
    TripletCLIP~\cite{patel2024tripletclip} & I & SLM (Mistral-7B) & General & E, F & F & - & - \\ 
    NaturalBench~\cite{li2024naturalbench} & I & Human-annotated & General & E, F & F & F & - \\
    VIOLIN~\cite{liu2020violin}             & V & Human-annotated & General & E, F & F & - & - \\ 
    VLContrastSet~\cite{park2022vlcontrastsets} & V & Rule + LLM & Specific (R) & E, F & F & F & - \\ 
    VFC~\cite{momeni2023verbsinaction}      & V & LLM (PaLM) & Specific (R) & E, F & F & F & - \\ 
    VideoCon~\cite{bansal2024videocon}      & V & LLM (PaLM-2) & Specific (R, S, A) & E, F & F & F & - \\
    Vinoground~\cite{zhang2024vinoground}   & V & Human + LLM & Specific (S) & E, F & F & F & - \\ 
    CompA~\cite{ghosh2024compa}             & A & LLM (GPT-4) & Specific (R, S) & E, F & F & F & - \\ 
    MATCH~\cite{kuan2025lalmhallucination} & A & Human-annotated & Specific (S) & E, F & F & F & - \\ 
    \midrule
    \textbf{MAC} (Ours) \hspace{130pt} & I, V, A & SLM (Llama3-8B) & General, Specific & E, F & E, F & E, F & E, F \\
    \bottomrule
    \end{tabular}
\end{adjustbox}
\end{center}
\caption{Overview of text-centric frameworks/benchmarks for multimodal compositionality. General/Specific denotes whether specific types of text operations are requested upon sample generation or not. 
Lexical indicates additional sample-wise constraints like instruction-following capability.
(\textbf{E}: Evaluate, \textbf{F}: Filter).
} 
\label{tab:overview}
\end{table*}

\section{Related Work}
\label{sec:relatedwork}

\textbf{Multimodal Compositional Reasoning}. 
Often studied in the vision-language domain, it refers to the structured relationship between words and their corresponding visual elements~\cite{thrush2022winoground}.
It serves as a key indicator of whether models truly understand multimodal contexts, impacting critical tasks such as negative sample mining~\cite{shekhar2017foil,zhao2022vlchecklist,yuksekgonul2022aro} and hallucination mitigation~\cite{li2023pope}.
To evaluate compositional reasoning, multiple benchmarks have been introduced to focus on robustness~\cite{park2024rococo}, systematicity~\cite{ma2023crepe}, and cross-domain alignment~\cite{yarom2023SeeTrue}.
Another line of work enhances compositional reasoning by curating training data~\cite{doveh2023SVLC,li2024llavascore,patel2024tripletclip} and regularizing learning objectives~\cite{oh2024FSCCLIP}. 
Recent efforts have expanded beyond image-text interactions to explore and improve compositionality in video-language~\cite{liu2020violin,park2022vlcontrastsets,momeni2023verbsinaction,bansal2024videocon} and audio-language contexts~\cite{ghosh2024compa}. 

Most closely related to our work is SugarCrepe~\cite{hsieh2023sugarcrepe}, which addresses the limitations of existing benchmarks 
by filtering nonsensical and non-fluent text to avoid trivial solutions.
NaturalBench~\cite{li2024naturalbench} focuses on generating challenging visual QA pairs 
easy for humans but difficult for models.
While both works employ adversarial filtering for compositional vulnerability, they primarily address bias balancing or human plausibility within image-text interactions.
In contrast, we approach compositionality from a modality-agnostic perspective and demonstrate this across image, video, and audio modalities.
While \citet{tang2024m3d} uses a claim manipulator model to contradict these modalities, our work highlights a key distinction by grounding the contradiction and diversity in a \textit{quantifiable} measure of deceiving the target multimodal representation.
Moreover, we extend our filtering criteria to better \textit{generate} such samples in terms of diversity and successful deception via self-training.


\noindent
\textbf{Multimodal Adversarial Attack on Text.}
Adversarial attacks~\cite{szegedy2013intriguing} manipulate input data to perturb a model's embedding space or induce incorrect predictions, systematically revealing vulnerabilities.
In continuous domains like images, attacks typically inject subtle noise to mislead inference or maliciously control model behavior~\cite{dong2018momentumattack,su2019pixelattack,shayegani2023jailbreak}.
In discrete domains like text, common strategies include identifying and replacing vulnerable words~\cite{li2020bertattack}, gradient-based attacks with Gumbel-softmax~\cite{guo2021gradient}, masked token perturbations~\cite{li2021clare}, and LLM-based refinement~\cite{mehrotra2023treeofattack}.

Text-based adversarial attacks can be extended to multimodal data, particularly targeting retrieval performance in image-text pairs by combining image noise injection and text perturbation.
For instance, Co-Attack~\cite{zhang2022CoAttack} applies multimodal distribution-aware collaborative perturbations to image-text pairs while maintaining crossmodal consistency.
Other methods enhance attack transferability via crossmodal guidance~\cite{lu2023SGA,xu2024DiffusionAttack,gao2024attackdiversification} or iterative search-based black-box attacks~\cite{yin2023vlattack,yu2023SparseMA}.
Recent studies have expanded attacks to video~\cite{yang2024VideoAttack} or audio~\cite{bagdasaryan2024adversarialillusion} beyond image-text pairs.
However, these approaches focus on embedding perturbations, often resulting in either simple paraphrasing or unnatural 
text modifications without considering their entailment with the original text.
To address these limitations, we instead apply a compositionality-aware modification that enables embedding-level perturbations while maintaining naturalness and semantic plausibility.

\begin{figure*}[t]
    \centering
    \includegraphics[width=\textwidth]{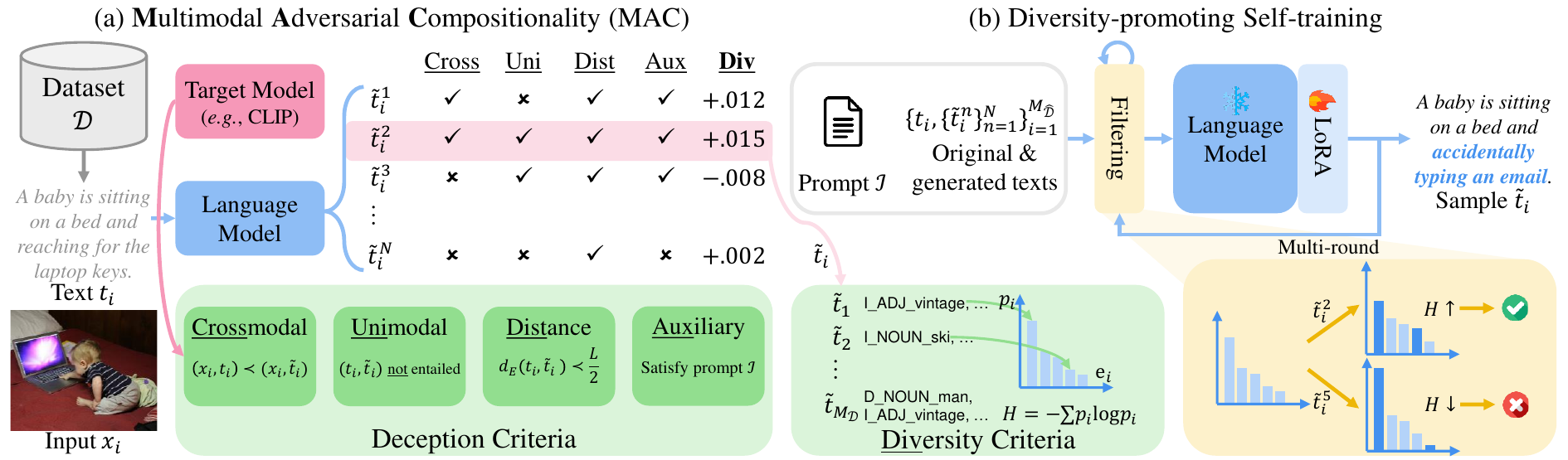}
    \caption{Overview of (a) multimodal adversarial compositionality and (b) diversity-promoting self-training.}
    \label{fig:benchmark}
\end{figure*}

\section{MAC: Multimodal Adversarial Compositionality}
\label{sec:benchmark}

\subsection{Problem Definition}
\label{subsec:mac_definition}

Our \textbf{M}ultimodal \textbf{A}dversarial \textbf{C}ompositionality benchmark (\textbf{MAC}) is illustrated in Fig.~\ref{fig:benchmark}.
Given a target pre-trained multimodal representation that we want to deceive (\textit{e.g.}, CLIP), MAC evaluates how effectively we can expose compositional vulnerabilities by updating text elements in multimodal data pairs.
We use text updates as an anchor 
since it allows for modality-agnostic assessment and is more intuitively aligned with human interpretation than noise injection~\cite{szegedy2013intriguing}.
Given a set of paired data $\mathcal{D}={(t_i, x_i)}_{i=1}^{M_\mathcal{D}}$, where $t_i$ represents text and $x_i$ represents a paired input modality (\textit{e.g.}, images), we aim to generate a set of adversarial text $\{\tilde{t}_i\}_{i=1}^{M_\mathcal{D}}$ that effectively exploit the compositional vulnerabilities 
of a target pre-trained multimodal representation  denoted by $f$, which
encodes both $t_i$ and $x_i$ into embeddings $y_{t_i},y_{x_i}=f(t_i,x_i) \in \mathbf{R}^d$.

The generation of adversarial text $\{\tilde{t}_i\}_{i=1}^{M_\mathcal{D}}$ comprises two key components: (1) an adversarial sample \textit{generator} $g$ that produces up to $N$ adversarial text samples $\{\tilde{t}_i^n\}_{n=1}^{N}$ under a specified budget constraint, and (2) a sample \textit{filterer} $h$ that identifies the most effective adversarial text sample $\tilde{t}_i$ from the $N$ candidates based on their potential to deceive the pre-trained model $f$.

Defining the multimodal compositionality problem as 
MAC offers several advantages.
First, since MAC does not assume a specific type of modality, it can be seamlessly applied to various formats including image, video, and audio. 
Second, previous compositionality frameworks that utilize rule-based or LLM-based generators for text updates, as well as our self-training-based generators (Sec.~\ref{sec:method}) can be consistently compared under our testbed to determine which framework more effectively deceives the target representation.

\subsection{Sample-wise Deception Evaluation}
\label{subsec:mac_deception}

\textbf{Crossmodal Criterion.}
First and foremost, the generated sample should achieve the intended attack.
The criterion is to deceive the target model $f$ such that the model determines the generated adversarial sample is more closely aligned with the input modality than the original text.
For an $i$-th data pair $(t_i, x_i)$ and a generated sample $\tilde{t}_i$, crossmodal attack success is
\begin{align}
    s_i^c &= \mathbf{I}(d_\theta(y_{t_i}, y_{x_i}) < d_\theta(y_{\tilde{t}_i}, y_{x_i})),
\label{eq:success_multimodal}
\end{align}
where $\mathbf{I}$ is an indicator function, and $d_\theta$ is an embedding distance, where we use cosine similarity.
For instance, in Fig.~\ref{fig:keyidea}-(c), 
$d_\theta(y_{t_i}, y_{x_i})$ and $d_\theta(y_{\tilde{t}_i}, y_{x_i})$ are 0.34 and 0.37, respectively, indicating a successful attack on CLIP.

\noindent
\textbf{Unimodal Criterion.}
While the crossmodal distance is a well-established measure, 
this criterion alone may lead to results that merely amount to paraphrasing, as demonstrated in various adversarial attack scenarios~\cite{zhang2022CoAttack,lu2023SGA}.
To prevent this, another crucial criterion is that there should be a meaningful semantic distinction between the generated sample and the original text.
Unimodal attack success for the $i$-th data pair is defined as follows:
\begin{align}
    s_i^u &= \Pi_j \mathbf{I}(l_j(t_i, \tilde{t}_i) < \tau),
\label{eq:success_unimodal}
\end{align}
where $\tau$ is a threshold for similarity and $l_j$ indicates an unimodal text model to measure entailment between two text samples~\cite{yarom2023SeeTrue,ma2023crepe}.
We use the agreement of multiple off-the-shelf NLI models~\cite{liu2019roberta,lewis2020bart,he2020deberta}. 
We use $\tau=0.5$, following~\citet{bansal2024videocon}.
In Fig.~\ref{fig:keyidea}-(c), all NLI models assess that the generated caption ``accidentally typing an email'' does not entail 
``reaching for the keys'', indicating a successful unimodal attack.
Note that we perform a preliminary evaluation using GPT-4 on 1K samples to verify the robustness of $s_i^u$, showing a concordance rate of over 93\% with GPT-4.

\noindent
\textbf{Distance Criterion.}
Model-based evaluation of unimodal gap effectively reflects the differences between embeddings; however, it may unfairly favor irrelevant text samples, which goes against the purpose of deceiving the original pair.
Therefore, 
the generated sample should 
execute attack
with only limited lexical deviation from the original sample:
\begin{align}
    s_i^d &= \mathbf{I}(d_E(t_i, \tilde{t}_i) < L_\mathcal{D}/2),
\label{eq:success_distance}
\end{align}
where $d_E$ is the Levenshtein distance between original and generated samples~\cite{ostrovsky2007low,andoni2020edit} 
and $L_\mathcal{D}$ is the average token length of dataset $\mathcal{D}$ for providing a dataset-specific limits in updates.
In Fig.~\ref{fig:keyidea}-(c), $d_E(t_i, \tilde{t}_i)=4$ is less than $L_\mathcal{D}/2 \approx 5.21$, satisfying the distance criterion.

\noindent
\textbf{Auxiliary Criterion.}
Lastly, we evaluate whether a generated sample follows a set of predefined rules.
For instance, as utilized by several frameworks in Table~\ref{tab:overview}, if generation should be performed through specific operations (\eg \texttt{swap}), failing to comply with this cannot be considered a successful deception.
Similarly, if trivial solutions are used, \textit{e.g.}, negation~\cite{ma2023crepe}, it is desirable for these to be filtered out as well.
The auxiliary attack success of $i$-th pair $s_i^a$ evaluates to true if it satisfies all predefined constraints (\textit{e.g.}, prompt) through rule-based lexical validation. 
In Fig.~\ref{fig:keyidea}-(b), the generated sample follows the \texttt{swap} operation by exchanging only two nouns (`baby' and `bed') without additional modifications. 

In total, the attack success rate $R$ is
\begin{align}
    R = \frac{1}{{M_\mathcal{D}}} \sum_i (s_i^c \cdot s_i^u \cdot s_i^d \cdot s_i^a).
\label{eq:asr}
\end{align}
Although these elements have been partially highlighted in previous research, our key contribution lies in bringing them together 
to quantify the attack effectiveness.
It enables consistent comparison across frameworks for revealing compositional vulnerabilities.

\subsection{Group-wise Diversity Evaluation}
\label{subsec:mac_diversity}

Another crucial criterion for successfully exposing compositional vulnerability is the diversity of generated samples.
While repeatedly employing similar and simple attack patterns might boost immediate attack success rates, such approaches are easily defensible and lack generalizability.
Indeed, 
when samples are generated without considering diversity,
the attack becomes overly focused on specific distributional weaknesses of the representation, resulting in frequently utilizing a limited set of vocabulary (\textit{e.g.}, man, woman, and vintage in Fig.~\ref{fig:app_tokendist} in Appendix~\ref{subsec:diversity_analysis}).
Therefore, a thorough analysis of pre-trained multimodal representation's compositional vulnerabilities necessitates the construction and utilization of adversarial samples that encompass diverse patterns of text updates, which has largely been overlooked.

To this end, we first construct a set of attribute-enriched tokens that represents a transformation from $t_i$ to $\tilde{t}_i$ through a series of insertion and deletion of words from the Levenshtein distance computation.
The token $e_i^j$ is defined as $\texttt{OP\_POS\_LEMMA}$, where $\texttt{OP, POS, LEMMA}$ corresponds to an ``word-level'' operation (insertion or deletion), a part-of-speech (POS) tag, and a lemmatized word, 
respectively (\textit{e.g.}, $\texttt{I\_NOUN\_man}$).
Such tokens distinguish which word-level operations or POS tags as well as words are involved when generating $\tilde{t}_i$ from $t_i$.

Using a set of attribute-enriched tokens from all data pairs, \textit{i.e.}, $\{\{e_i^j\}_{j=1}^{E_i}\}_{i=1}^{M_\mathcal{D}}$, we compute probability distribution of unique tokens with respect to their frequency to obtain entropy $H=-\sum_j p_j \log p_j$, which indicates the extent to which the distribution is spread across different tokens.
$p_j$ denotes the probability of a $j$-th unique token and $E_i$ is the number of tokens for an $i$-th sample.
Note that higher $H$ implies a more diverse set of lexical operations are involved when composing deceptive samples.
To prevent pathological cases where the generator might produce arbitrary text to achieve high entropy values, we only consider samples that meet the edit distance criterion (Eq.~\ref{eq:success_distance}) for diversity evaluation, discarding attribute-enriched tokens from samples that exceed this threshold. This ensures that our diversity metrics reflect meaningful variations in text transformations rather than random deviations from the ground truth.

Since $H$ does not account for how many unique tokens are involved in generation, we also report two additional complementary measures.
Following \citet{li2016diversity} and \citet{zhang2021trading}, distinct-1 ($D_1$) captures the ratio of unique tokens out of all tokens.
On the other hand, the normalized entropy $\hat{H}$ compromises $H$ and $D_1$ by normalizing $H$ by the number of unique tokens.

\subsection{Threat Model Categorization}
\label{subsec:mac_threatmodel}

In a nutshell, we can categorize the threat model of our framework by following the taxonomy established in adversarial learning~\cite{zhang2020adversarial,laidlaw2021perceptual,schwinn2023adversarial,shayegani2023survey,vassilev2024adversarial}:

\begin{itemize}
\setlength{\itemsep}{0pt}
    \item \textbf{Model knowledge} - (i) \textit{Gray-box} for crossmodal assessment (\textit{e.g.}, CLIP, LanguageBind); we use only output embeddings with respect to queries without accessing gradients and model parameters. (ii) Black-box for unimodal assessment; we use entailment scores of off-the-shelf NLI models without other information.
    \item \textbf{Attack target} - Untargeted; we induce incorrect predictions instead of eliciting specific responses.
    \item \textbf{Attack granularity} - Mix of word-level and sentence-level perturbation
    \item \textbf{Perturbation constraint} - Distance and auxiliary criteria (\S\ref{subsec:mac_deception}) and diversity evaluation (\S\ref{subsec:mac_diversity}) for perceptually plausible attacks
    \item \textbf{Evaluation} - The sample-wise attack success rate and group-wise diversity evaluation
    \item \textbf{Modality} - Language + X, where X can be image, video, and audio
    \item \textbf{Budget} - Number of sampling with LLM ($N$), which will be further discussed (\S\ref{sec:method}).
\end{itemize}


\section{Approach} 
\label{sec:method}

\subsection{Motivation}
Among diverse generators $g$ (\eg rule-based, human-based, LLM-based) in Table~\ref{tab:overview}, we prioritize LLM-based methods for the following reasons: (1) Rule-based methods (\eg word swapping) often produce nonsensical and non-fluent text. Additionally, these methods tend to yield simplistic text focused on specific scenarios that models can easily defend against~\cite{hsieh2023sugarcrepe}.
(2) While human-generated annotations provide fluent text, they are difficult to scale due to resource constraints and the labor-intensive nature of the annotation process.
(3) LLMs address these limitations by generating fluent text at scale. Thanks to these advantages, recent multimodal compositionality studies have increasingly adopted LLM-based methods instead of relying on rule-based or human-annotated methods.

\subsection{Preliminary: Revealing Compositional Vulnerabilities via Filtering}
\label{subsec:filtering}

While attacks in vision-language compositionality literature typically occur only once ($N=1$), leveraging multiple attempts ($N>1$) with sample selection could be more effective in revealing such vulnerabilities~\cite{shekhar2017foil, yarom2023SeeTrue, park2022vlcontrastsets}.
To incorporate sample selection into MAC, we adopt a Best-of-$N$ strategy—a widely used and general sampling approach—that selects the best sample.
Given $N$ samples $\{\tilde{t}_i^n\}_{n=1}^N$, it prioritizes those that meet all sample-wise criteria in Sec.~\ref{subsec:mac_deception}. 
If such samples exist, we randomly select from them; otherwise, we sample randomly from the entire set:
\begin{align}
    \mathcal{T}_{i} &= \{ \tilde{t}_i^n \mid (s_i^c \cdot s_i^u \cdot s_i^d \cdot s_i^a)(\tilde{t}_i^n, t_i, x_i) = 1 \}, \\
    \tilde{t}_i &\sim 
    \begin{cases}
        \mathrm{Uniform}(\mathcal{T}_{i}), & \text{if } \mathcal{T}_{i} \neq \emptyset, \\
        \mathrm{Uniform}(\{\tilde{t}_i^n\}_{n=1}^N), & \text{otherwise}.
    \end{cases} 
\label{eq:filtering}
\end{align}

As demonstrated in Table~\ref{tab:main_result_table}, while the filtering approach with $N>1$ shows improved performance compared to baseline methods, this approach faces several limitations.
First, the computational cost scales linearly with $N$ when generating samples for each pair, and the time complexity increases significantly when performed sequentially (see Table~\ref{tab:analysis_TTC_selection} in Appendix~\ref{subsec:mac_across_ttc}).
Moreover, relying on larger $N$ masks the true effectiveness of adversarial strategies by enabling brute-force attempts.
Thus, we limit $N$ to evaluate attack efficiency rather than persistence.

\subsection{Self-training}

To address the limitations of filtering-based approaches, we propose a learnable method designed to enhance the exposure of compositional vulnerabilities for the first time.
Given the absence of annotations or ground truth, we employ self-training~\cite{huang2023selfimprove} by promoting responses similar to the condition-satisfying samples generated by the base language model.
This approach falls into the category of rejection sampling fine-tuning (RFT)~\cite{touvron2023llama2}.
From the training set $\mathcal{D_\text{train}}={(t_i, x_i)}_{i=1}^{M_\mathcal{D_\text{train}}}$, we first generate and filter samples $\{\tilde{t}_i\}_{i=1}^{M_\mathcal{D_\text{train}}}$ using Eq.~\ref{eq:filtering}, then only use $M_{\hat{\mathcal{D}}}$ successful adversarial samples to train the model using RFT loss:
\begin{align}
\{\tilde{t}_i\}_{i=1}^{M_{\hat{\mathcal{D}}}} &= \left\{\tilde{t}_i \mid s_i^c \cdot s_i^u \cdot s_i^d \cdot s_i^a = 1 \right\}, 
\label{eq:labelselection}
\end{align}
\begin{align}
\mathcal{L} &= -\frac{1}{M_{\hat{\mathcal{D}}}} \sum_{i} \sum_j \log g(\tilde{t}_{i,j} | \tilde{t}_{i,<j}, \mathcal{I}, t_i; \Theta),
\label{eq:rftloss}
\end{align}
where $\mathcal{I}$ denotes instruction prompt and $\Theta$ is a set of learnable parameters of the generator $g$.

As shown in Table~\ref{tab:main_result_table}, self-training significantly improves the attack success rate by learning to favor samples that effectively attack vulnerabilities with small $N$ (\eg $N=4$).
To further enhance attack performance beyond na\"ive self-training, one can either train with a larger $N(>4)$ or iterate self-training 
as needed. While self-training requires additional computational cost, it can be amortized during inference and leads to more efficient inference by reducing the number of attempts $N$ required to achieve high attack success rates. 
In our experiments, we set $N=64$ as the default value for large-$N$ distilled self-training.

\setlength{\textfloatsep}{5pt}
\begin{algorithm}[t]
\caption{Diversity-promoting Self-training Data Selection}
\label{alg:diversity}
\begin{algorithmic}
\REQUIRE Set of $N$ samples $\{\tilde{t}_i^n\}_{n=1}^N$ generated for each training instance $i \in [1,M_{\hat{\mathcal{D}}}]$, and diversity function $H$
\ENSURE Diverse successful samples $\{\tilde{t}_i\}_{i=1}^{M_{\hat{\mathcal{D}}}}$
\STATE Initialize $\{\tilde{t}_i\}_{i=1}^{M_{\hat{\mathcal{D}}}}$ randomly from $\{\tilde{t}_i^n | (s_i^c \cdot s_i^u \cdot s_i^d \cdot s_i^a)(\tilde{t}_i^n, t_i, x_i) = 1\}$ 
\FOR{iteration $k = 1$ to $K$}
    \FOR{$i = 1$ to $M_{\hat{\mathcal{D}}}$}
        \STATE $\mathcal{T}_i \gets \{\tilde{t}_i^n | (s_i^c \cdot s_i^u \cdot s_i^d \cdot s_i^a)(\tilde{t}_i^n, t_i, x_i) = 1\}$
        \STATE $\tilde{t}_i \gets \mathrm{argmax}_{\tilde{t}^n_i \in \mathcal{T}_i} H(\tilde{t}_1, ..., \tilde{t}^n_i, ..., \tilde{t}_{M_{\hat{\mathcal{D}}}})$
    \ENDFOR
\ENDFOR
\RETURN $\{\tilde{t}_i\}_{i=1}^{M_{\hat{\mathcal{D}}}}$
\end{algorithmic}
\end{algorithm}

\subsection{Diversity-promoting Self-training}
\label{subsec:training}

Although effective at generating successful attacks, self-training tends to generate monotonous samples focused on specific distributional weaknesses rather than maintaining sample diversity, resulting in decreased diversity.
The selection of samples involved in training is therefore more important than the training process itself from the perspective of exposing compositional vulnerability.
To enhance diversity while maintaining successful attacks, we introduce a Gibbs sampling-based selection process described in Algorithm~\ref{alg:diversity}.
This approach iteratively selects sample that maximize diversity among successful attacks.
While we employ entropy $H$ as a representative diversity metric, it can be substituted with any quantifiable diversity measure (e.g., $D_1$).


\section{Experiments}
\label{sec:experiments}

\begin{table*}[t!]
\centering
\begin{adjustbox}{width=\linewidth}
\begin{tabular}{lcccccccccccc}
\toprule
\multirow{3}{*}{\makecell[l]{Method}}
& \multicolumn{4}{c}{\textbf{(a) Image (CLIP/COCO)}} & \multicolumn{4}{c}{\textbf{(b) Video (LB/MSRVTT)}} & \multicolumn{4}{c}{\textbf{(c) Audio (LB/AudioCaps)}} \\
& \multicolumn{2}{c}{ASR$_\uparrow$} & \multicolumn{2}{c}{Diversity$_\uparrow$} & \multicolumn{2}{c}{ASR$_\uparrow$} & \multicolumn{2}{c}{Diversity$_\uparrow$} & \multicolumn{2}{c}{ASR$_\uparrow$} & \multicolumn{2}{c}{Diversity$_\uparrow$} \\
\cmidrule(r{0.3em}){2-3} \cmidrule(r{0.3em}){4-5} \cmidrule(r{0.3em}){6-7} \cmidrule(r{0.3em}){8-9} \cmidrule(r{0.3em}){10-11} \cmidrule(r{0.3em}){12-13}
& Cross & Total & $H$ & $D_1$
& Cross & Total & $H$ & $D_1$
& Cross & Total & $H$ & $D_1$ \\
\midrule
\rowcolor{gray!15}
\multicolumn{13}{l}{\textbf{N=1}} \\
RoCOCO$_{\text{rand-voca}}$~\citep{park2024rococo} & 24.33 & 1.99 & \underline{7.642} & \textbf{0.196} & - & - & - & - & - & - & - & - \\
RoCOCO$_{\text{Danger}}$~\citep{park2024rococo} & 20.24 & 7.88 & 4.454 & 0.052 & - & - & - & - & - & - & - & - \\
RoCOCO$_{\text{same-concept}}$~\citep{park2024rococo} & 17.09 & 5.29& 7.098 & 0.088 & - & - & - & - & - & - & - & - \\
RoCOCO$_{\text{diff-concept}}$~\citep{park2024rococo} & 17.92 & 2.75 & 7.128 & 0.089 & - & - & - & - & - & - & - & - \\
SugarCrepe$^\ast$~\citep{hsieh2023sugarcrepe} & 10.84 & 2.40 & 7.312 & 0.103 & - & - & - & - & - & - & - & - \\
LLaVa-Score$^\ast$~\citep{li2024llavascore} & 24.81 & 5.71 & 7.201 & 0.110 & - & - & - & - & - & - & - & - \\
TripletCLIP~\citep{patel2024tripletclip} & 12.81 & 6.34 & 7.551 & 0.092 & - & - & - & - & - & - & - & - \\
VideoCon$^\ast$~\citep{bansal2024videocon} & - & - & - & - & 16.30 & 7.10 & 6.702 & 0.610 & - & - & - & - \\
Deceptive-General Prompt (zero-shot) & 28.52 & 6.88 & 7.562 & \underline{0.131} & 32.20 & 7.70 & 6.809 & \underline{0.638} & 28.68 & 10.47 & \underline{6.572} & \underline{0.182} \\
\midrule
\rowcolor{gray!15}
\multicolumn{13}{l}{\textbf{N=4}} \\
SeeTrue~\citep{yarom2023SeeTrue} & 34.67 & 23.33 & 7.168 & 0.124 & - & - & - & - & - & - & - & - \\
VFC$^\ast$~\citep{momeni2023verbsinaction} & - & - & - & - & 42.60 & 36.90 & 5.929 & 0.381 & - & - & - & - \\
CompA$^\ast$~\citep{ghosh2024compa} & - & - & - & - & - & - & - & - & 49.38$^\dagger$ & 5.76$^\dagger$ & 6.009$^\dagger$ & 0.171$^\dagger$ \\
Deceptive-General Prompt (zero-shot) & 37.29 & 19.19 & 7.571 & 0.130 & 42.40 & 24.80 & 6.808 & 0.626 & 42.60 & 29.02 & 6.566 & 0.172 \\
+ Self-Train & 43.08 & 34.64 & 7.507 & 0.120 & 48.90 & 39.70 & \underline{6.900} & 0.587 & 55.37 & 47.35 & 6.472 & 0.157 \\
+ Self-Train + Large-$N$ Distilled & \textbf{48.29} & \underline{42.03} & 7.452 & 0.117 & \underline{52.90} & \underline{44.20} & 6.839 & 0.594 & \underline{58.38} & \underline{51.57} & 6.508 & 0.157 \\
+ Self-Train + Large-$N$ Distilled + Diversity-Promoted (\textbf{Ours}) & \underline{47.93} & \textbf{42.10} & \textbf{7.747} & 0.129 & \textbf{53.50} & \textbf{45.60} & \textbf{7.125} & \textbf{0.667} & \textbf{60.25} & \textbf{52.87} & \textbf{6.868} & \textbf{0.191} \\
\bottomrule
\end{tabular}
\end{adjustbox}
\caption{
Main Results. 
`-' indicates that the method is not applicable.
($^\ast$: the prompts from the original papers are slightly modified. 
$^\dagger$: the results are computed for a subset to which the method can be applied).
}
\label{tab:main_result_table}
\end{table*}

\begin{table}[t!] \begin{center}
    \small
   \begin{tabular}{lcccc}
    \toprule
ASR$_\text{Total}$& \;\;CLIP\;\; & \;SigLIP\; & NegCLIP & \;\;BLIP\;\; \\
    \midrule
CLIP 
 & \makecell[c]{42.10\\\tiny{(+22.91)}}
 & \makecell[c]{28.63\\\tiny{(+15.68)}}
 & \makecell[c]{24.84\\\tiny{(+12.71)}}
 & \makecell[c]{25.25\\\tiny{(+14.13)}}\\
SigLIP
 & \makecell[c]{29.37\\\tiny{(+16.13)}}
 & \makecell[c]{41.04\\\tiny{(+21.32)}}
 & \makecell[c]{23.84\\\tiny{(+12.17)}}
 & \makecell[c]{25.01\\\tiny{(+13.76)}}\\
NegCLIP
 & \makecell[c]{25.40\\\tiny{(+12.68)}}
 & \makecell[c]{23.63\\\tiny{(+11.47)}}
 & \makecell[c]{40.81\\\tiny{(+20.10)}}
 & \makecell[c]{23.77\\\tiny{(+12.33)}}\\
BLIP
 & \makecell[c]{19.84\\\tiny{(+10.60)}}
 & \makecell[c]{19.11\\\tiny{(+10.04)}}
 & \makecell[c]{18.02\\\tiny{(+8.94)}}
 & \makecell[c]{32.50\\\tiny{(+17.80)}}\\
    \bottomrule
   \end{tabular}
   \caption{Cross-model transfer analysis ($N=4$). Columns are source models for filtering, and rows are target models for evaluation. Numbers in parentheses are absolute gains from our proposed self-training compared to the zero-shot baselines.}
   \label{tab:analysis_clip_transfer}
\end{center}
\end{table}

\subsection{Evaluation Protocol}
\label{subsec:evaluation_protocol}

\textbf{Target representation.}
We primarily use CLIP~\cite{radford2021clip} and LanguageBind (LB)~\cite{zhu2023languagebind} as target multimodal representations. They are representative models with dual-modality and multi-modality pre-training.
Additionally, to analyze the transferability of deception across different representations, we also evaluate SigLIP~\cite{zhai2023siglip}, NegCLIP~\cite{yuksekgonul2022aro}, and BLIP~\cite{li2022blip}.

\noindent
\textbf{Sample generation.}
Our methodology operates by modifying text (Sec.~\ref{subsec:mac_definition}).
We generate samples that reveal compositional vulnerability using representative multimodal datasets: COCO~\cite{lin2014coco} for image, MSRVTT~\cite{xu2016msrvtt} for video, and AudioCaps~\cite{kim2019audiocaps} for audio.

Unless mentioned otherwise, we use Llama-3.1-8B~\cite{dubey2024llama3} 
for sample generation and self-training.
We explore its applicability across different LLMs, including GPT-4o~\cite{achiam2023gpt4}, noting that larger or proprietary models do not necessarily lead to more effective deception, as discussed in Appendix~\ref{subsec:mac_across_llms}.
We employ two instruction prompts (\textit{i.e.}, $\mathcal{I}$ in Eq.~\ref{eq:rftloss}).
The \textit{deceptive-general prompt} instructs to expose vulnerability without constraints on text updates, while the \textit{deceptive-specific prompt} instructs to perform text updates corresponding to \texttt{replace, swap}, and \texttt{add} based on taxonomy from existing literature, as in Table~\ref{tab:overview}.
See Appendix~\ref{subsec:prompt_demonstration} for prompt demonstrations.
For better performance, we primarily use the general prompt.

\noindent
\textbf{Evaluation metrics.}
We conduct sample-wise and group-wise evaluations as described in Sec.~\ref{sec:benchmark}. For sample-wise evaluation, we report the attack success rate (ASR) focusing on crossmodal criterion (\textit{Cross}) and all criteria (\textit{Total}), while for group-wise diversity evaluation, we  report entropy ($H$) and distinct-1 ($D_1$). 
Fine-grained performance comparisons 
are discussed in Appendix~\ref{subsec:ablation_study}.

\noindent
\textbf{Baselines.}
We establish a set of competitive baselines using existing compositionality frameworks.
For models generating with $N=1$ budget, we utilize RoCOCO~\cite{park2024rococo}, SugarCrepe~\cite{hsieh2023sugarcrepe}, LLaVa-Score~\cite{li2024llavascore}, TripletClip~\cite{patel2024tripletclip}, and VideoCon~\cite{bansal2024videocon}.
For filtering-based models, we employ SeeTrue~\cite{yarom2023SeeTrue}, VFC~\cite{momeni2023verbsinaction}, and CompA~\cite{ghosh2024compa}, using $N=4$ for inference. 
For the studies that use proprietary models like GPT-4, we substitute Llama-3.1-8B for it and modify the prompts to ensure effective sample generation with this model for fair comparison and cost constraints.
For experimental details, see Appendix~\ref{sec:experimental_details}.

\subsection{Experimental Results}
\label{subsec:experimental_results}

Table~\ref{tab:main_result_table} summarizes the overall results, showing our approach outperforms prior methods in both ASR and diversity.
As evident from RoCOCO’s first two variants, there exists a trade-off where maximizing ASR leads to a sharp decline in diversity and vice versa, indicating that focusing on either metric alone is far from optimal.
Generating multiple samples and applying filtering improves ASR across all modalities compared to $N=1$, though this does not translate to enhanced diversity.
See Appendix~\ref{subsec:diversity_analysis} (Fig.~\ref{fig:app_tokendist}) for qualitative distribution in terms of diversity.

\begin{figure*}[t]
    \centering
    \includegraphics[width=\textwidth]{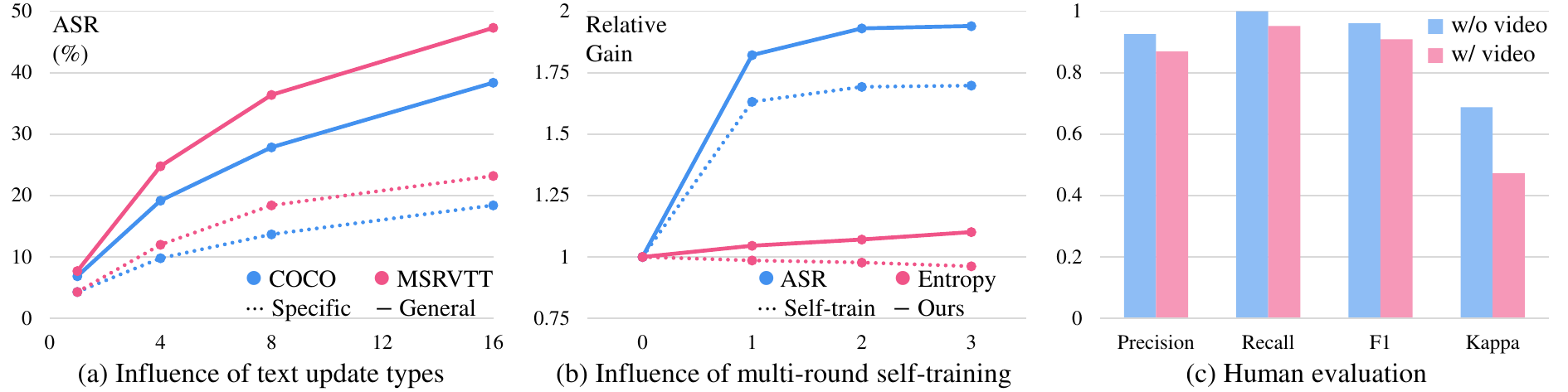}
    \caption{Analysis of our proposed framework. Please refer to Sec.~\ref{subsec:performance_analysis} for detailed explanation.}
    \label{fig:analysis}
\end{figure*}

\begin{figure}[t]
    \centering
    \includegraphics[width=0.49\textwidth]{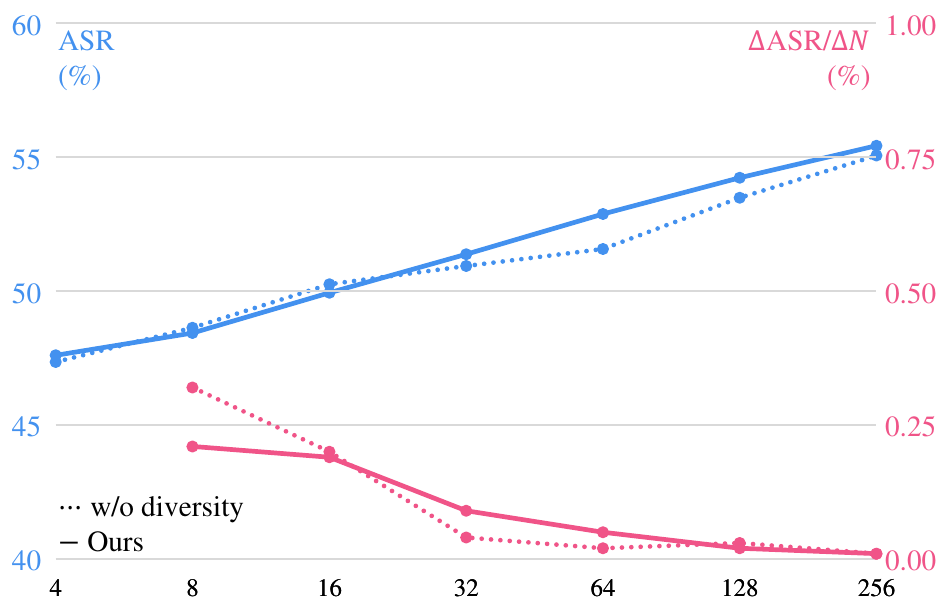}
    \caption{Influence of $N$ in self-training.}
    \label{fig:analysis_rebuttal}
    \vspace{0pt}
\end{figure}

\begin{figure}[t]
    \centering
    \includegraphics[width=0.49\textwidth]{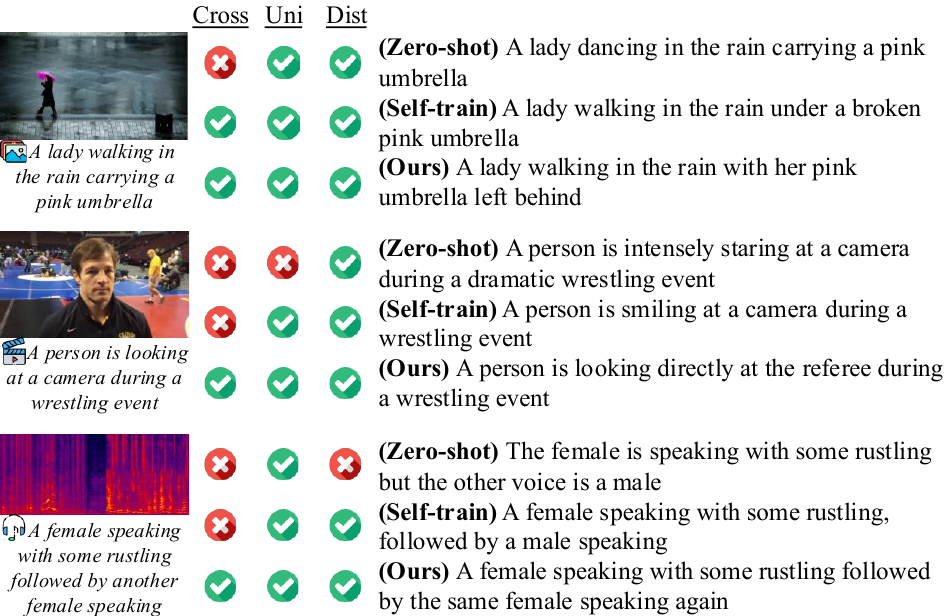}
    \caption{
Qualitative examples from COCO, MSRVTT, and AudioCaps datasets (from top to bottom).
}
    \label{fig:qual}
\end{figure}

The last four rows reveal the ablation study of our method.
Using only the deceptive-general prompt yields performance comparable to existing methods.
Adding self-training for a single iteration dramatically increases ASR, \textit{i.e.}, +68\% on average, 
underscoring its role in addressing compositionality.
Yet, this alone does not enhance diversity and may even reduce it. 
This implies na\"ive self-training, while effective for ASR, falls short in diverse exposure of compositional vulnerability.
Instead, incorporating diversity-promoting filtering leads to consistent improvements in both 
diversity metrics
without sacrificing ASR (+2\%), advancing the pareto front in the attack-diversity trade-off.

Table~\ref{tab:analysis_clip_transfer} examines the transferability of deceptive samples across multimodal representations. 
The results show high transferability, 
often exceeding the best performing baseline (23.33).
Notably, the performance gains from self-training are substantial across all settings, achieving 2.1$\times$ improvement on average. 
BLIP shows slightly lower performance presumably due to its use of yes/no classification logits instead of embedding similarity.


\subsection{Performance Analysis}
\label{subsec:performance_analysis}

\textbf{General vs. specific prompt.}
As summarized in Table~\ref{tab:overview}, various compositionality frameworks employ either general or specific types of prompts, necessitating an analysis of their effectiveness in ASR. 
Fig.~\ref{fig:analysis}-(a) compares performance under different instruction types for generation budget $N$. 
Methods without specific text update constraints consistently outperform constrained ones, with this trend persisting as $N$ increases. 
Notably, our self-training approach with $N=4$ matches the performance of non-self-training methods with an $N=16$ budget.

\noindent
\textbf{Influence of multi-round self-training.}
Self-training enables multiple iterations by refining filtering models across training rounds.
Fig.~\ref{fig:analysis}-(b) shows the relative gains of diversity-promoting vs. na\"ive self-training on AudioCaps.
Our self-training significantly improves ASR performance, reaching saturation by the third round.
While entropy degrades with conventional self-training, our approach sustains continuous improvement.
For MSRVTT results, please refer to Appendix~\ref{subsec:multi_round_self_train}.

\noindent
\textbf{Influence of large $N$ in self-training.}
To better understand the influence of $N$ in distillation-based self-training, we report the ASR of our method using AudioCaps in Fig.~\ref{fig:analysis_rebuttal}.
While increasing $N$ does not display a clear signal of saturation, the relative performance gain with respect to $N$ ($\Delta \mathrm{ASR}/\Delta N$) does.
This diminishing return suggests that $N=64$ offers a reasonable balance between performance improvement and time constraint.

\noindent
\textbf{Human evaluation.}
A potential limitation is our reliance on the model-based unimodal entailment assessment, necessitating evaluation on human agreement.
Fig.~\ref{fig:analysis}-(c) compares our criterion against human evaluation by five annotators on 50 random MSRVTT test samples. 
Results show high agreement (F1 > 0.9) regardless of video presence, with moderate to substantial inter-annotator agreement $\kappa$~\cite{fleiss1971kappa}.
Although $\kappa$ is slightly lower for evaluations with videos—likely due to subjective interpretation of longer contexts—overall agreement remains strong (F1 = 0.9091), confirming the reliability of our unimodal assessment.

\noindent
\textbf{Qualitative examples.}
Fig.~\ref{fig:qual} compares generated samples from variants of our method across different modalities.
Compared to other variants, our self-training successfully applies various modification without being constrained to specific patterns.
Additional examples are provided in Appendix~\ref{subsec:qualitative_results}.

\section{Conclusion}
\label{sec:conclusion}
We explored the compositional vulnerability of pre-trained multimodal representations using LLMs.
First, we established a testbed by proposing \textbf{MAC}, which provides a comprehensive set of criteria for evaluating how effectively and diversely a target representation can be deceived.
Furthermore, we suggested the application of self-training to multimodal compositionality for the first time via iterative RFT with diversity-promoting filtering to improve both ASR and diversity.
Lastly, our modality-agnostic assessment allowed for a thorough analysis of compositional vulnerabilities across image, video, and audio modalities, where our method consistently outperformed prior arts across various target representations.
Our benchmark’s modality-agnostic design opens avenues for extending vulnerability analysis to less-explored modalities like IMU or tactile sensing, even in the absence of multimodal LLMs capable of processing these data types.

\section*{Limitations}
Our work focused on short captions in exploring multimodal adversarial compositionality.
Extending MAC (\textit{i.e.}, deceiving pre-trained multimodal representations) to longer, detailed captions~\cite{onoe2024docci,chen2024sharegpt} represents a distinct but promising research direction, as it would require more sophisticated attack strategies that consider long-range dependencies and contextual relationships throughout the caption to successfully deceive target representations.

\section*{Ethics Statement}
Since our work uses language models to generate adversarial captions to reveal compositional vulnerabilities, they might potentially generate biased or toxic content. 
We encourage practitioners who wish to use generated captions to carefully monitor and filter outputs to prevent unintended harmful content.

For human evaluation, we worked with annotators primarily from the US, UK, Canada, New Zealand, and Australia, ensuring fair compensation above their local minimum wages (averaging \$18 per hour).
Please refer to Appendix~\ref{subsec:human_eval} for details.

\section*{Acknowledgments}
This work was supported by the
Institute of Information \& Communications Technology Planning \& Evaluation (IITP) grant (No.~RS-2019-II191082, RS-2021-II211343, No.~RS-2022-II220156),
the National Research Foundation of Korea (NRF) grant (No.~2023R1A2C2005573),
and the IITP-ITRC (Information Technology Research Center) grant (IITP-2025-RS-2024-00437633) funded by the Korea government (MSIT).
Gunhee Kim is the corresponding author.

\bibliography{paper-acl25-advcomp}

\clearpage
\appendix

\section{Experimental Details}
\label{sec:experimental_details}

\subsection{Dataset}
We used standard train and test sets commonly employed in multimodal retrieval tasks as follows.

For COCO~\cite{lin2014coco}, we adopt the Karpathy test split~\cite{karpathy2017imagecap} as the test set, which consists of 5,000 images paired with 25,010 captions. The train set corresponds to the COCO 2014 train split, containing 83,287 images and 414,113 captions.
For MSRVTT~\cite{xu2016msrvtt}, we utilize the MSRVTT 1K-A split~\cite{yu2018jsfusion} as the test set, which includes 1,000 videos, each associated with a single caption. The train set corresponds to the MSRVTT 9K train split, containing 9,000 videos with 180,000 captions.
For AudioCaps~\cite{kim2019audiocaps}, we use the test split from~\citet{oncescu2021audioretrieval}, which consists of 816 audio clips with 4,080 captions. The train set corresponds to the train split from~\citet{oncescu2021audioretrieval}, which includes 49,291 audio clips, each paired with a single caption.
All datasets contain English language captions and are publicly available, used in accordance with their respective licenses for research purposes.

Note that each train set $(x_i, t_i)$ does not include a label for deceptive caption supervision. This absence of supervision serves as the primary motivation for our self-training approach, which aims to generate deceptive captions $\tilde{t}_i$.

\subsection{Models}
\textbf{Target models.}  
For target pre-trained multimodal representations for evaluating crossmodal criterion in Sec.~\ref{subsec:mac_deception}, we utilize:  
CLIP\footnote{\href{https://huggingface.co/laion/CLIP-ViT-H-14-laion2B-s32B-b79K}{laion/CLIP-ViT-H-14-laion2B-s32B-b79K}},  
SigLIP\footnote{\href{https://huggingface.co/google/siglip-so400m-patch14-384}{google/siglip-so400m-patch14-384}},
NegCLIP\footnote{\href{https://github.com/mertyg/vision-language-models-are-bows}{https://github.com/mertyg/vision-language-models-are-bows}}
BLIP\footnote{\href{https://huggingface.co/Salesforce/blip-itm-base-coco}{Salesforce/blip-itm-base-coco}},  
LanguageBind$_\text{Video}$\footnote{\href{https://huggingface.co/LanguageBind/LanguageBind_Video_FT}{LanguageBind/LanguageBind\_Video\_FT}},  
and LanguageBind$_\text{Audio}$\footnote{\href{https://huggingface.co/LanguageBind/LanguageBind_Audio_FT}{LanguageBind/LanguageBind\_Audio\_FT}}.

\textbf{NLI models.}  
For NLI models for evaluating the unimodal criterion in Sec.~\ref{subsec:mac_deception}, we utilize:  
RoBERTa\footnote{\href{https://huggingface.co/FacebookAI/roberta-large-mnli}{FacebookAI/roberta-large-mnli}},  
DeBERTa\footnote{\href{https://huggingface.co/microsoft/deberta-xlarge-mnli}{microsoft/deberta-xlarge-mnli}},  
and BART\footnote{\href{https://huggingface.co/facebook/bart-large-mnli}{facebook/bart-large-mnli}}.


\textbf{LLMs.}  
For LLMs, we use:  
Llama-3.1-8B\footnote{\href{https://huggingface.co/meta-llama/Meta-Llama-3.1-8B-Instruct}{meta-llama/Meta-Llama-3.1-8B-Instruct}},  
Llama-3.1-70B (Q4\_0)\footnote{\href{https://ollama.com/library/llama3.1:70b-instruct-q4_0}{Ollama Llama-3.1-70B (Q4\_0)}},  
Qwen-2.5-7B\footnote{\href{https://huggingface.co/Qwen/Qwen2.5-7B-Instruct}{Qwen/Qwen2.5-7B-Instruct}},
Gemma-2-9B\footnote{\href{https://huggingface.co/google/gemma-2-9b-it}{google/gemma-2-9b-it}},
and GPT-4o$_\texttt{2024-08-06}$.
Here, Q4\_0 denotes a 4-bit quantized version of the model.

\subsection{Prompt Demonstration}  
\label{subsec:prompt_demonstration}
\textbf{Deceptive-General Prompt.}  
The deceptive-general prompt is presented in Table~\ref{tab:deceptive_general_prompt}.  

\textbf{Deceptive-Specific Prompt.}  
The deceptive-specific prompts, tailored for different modification types, are presented as follows:  
\begin{itemize}  
    \item \textbf{Replacement Prompts:}  
    \begin{itemize}  
        \item Table~\ref{tab:deceptive_specific_prompt_replace_object}: Replacing objects.  
        \item Table~\ref{tab:deceptive_specific_prompt_replace_attribute}: Replacing attributes.  
        \item Table~\ref{tab:deceptive_specific_prompt_replace_relation}: Replacing relationships.  
        \item Table~\ref{tab:deceptive_specific_prompt_replace_count}: Replacing numerical counts.  
    \end{itemize}  
    \item \textbf{Addition Prompts:}  
    \begin{itemize}  
        \item Table~\ref{tab:deceptive_specific_prompt_add_object}: Adding objects.  
        \item Table~\ref{tab:deceptive_specific_prompt_add_attribute}: Adding attributes.  
    \end{itemize}  
    \item \textbf{Swap Prompts:}  
    \begin{itemize}  
        \item Table~\ref{tab:deceptive_specific_prompt_swap_object}: Swapping objects.  
        \item Table~\ref{tab:deceptive_specific_prompt_swap_attribute}: Swapping attributes.  
    \end{itemize}  
\end{itemize}

\begin{table}[htbp]
\scriptsize
\centering
\begin{tabular}{@{}p{\linewidth}@{}}
\toprule
\textbf{Deceptive-General Prompt}\\
\midrule
You will be given a caption describing the \{contents\_modality\}. Your task is to generate a hard negative caption using the criteria below:

~

***

[Generation Criteria]

1. Ensure the new caption has higher similarity to the \{contents\_modality\} in \{contents\_modality\}-text crossmodal model than the given caption.

2. Introduce a contradiction compared to the given caption, but avoid simple negations (e.g., using words like "no", "not", "empty", or "without").

3. Make fewer than \{max\_word\_distance\_plus\_one\} word-level changes (add, delete, or substitute words) to the given caption without fully rewriting it to generate the new caption.

~

[Given Caption]

- \{caption\}

***

~

Write only the new caption starting with "Generated Caption: ", without explanation. \\
\bottomrule
\end{tabular}
    \caption{Deceptive-general prompt.}
    \label{tab:deceptive_general_prompt}
\end{table}

\begin{table}[htbp]
\scriptsize
\centering
\begin{tabular}{@{}p{\linewidth}@{}}
\toprule
\textbf{Deceptive-Specific Prompt (\texttt{replace-object})}\\
\midrule
You will be given a caption describing the \{contents\_modality\}. Your task is to generate a hard negative caption based on the "object replacement" scenario using the criteria below:

~

***

[Generation Criteria]

1. Replace a key object in the given caption with a new object that is not in the given caption.

2. Ensure the new caption has higher similarity to the \{contents\_modality\} in \{contents\_modality\}-text crossmodal model than the given caption.

3. Introduce a contradiction compared to the given caption, but avoid simple negations (e.g., using words like "no", "not", "empty", or "without").

4. Make fewer than \{max\_word\_distance\_plus\_one\} word-level changes (add, delete, or substitute words) to the given caption without fully rewriting it to generate the new caption.

~

[Given Caption]

- \{caption\}

***

~

Write only the new caption starting with "Generated Caption: ", without explanation. \\
\bottomrule
\end{tabular}
    \caption{Deceptive-specific prompt (\texttt{replace-object}).}
    \label{tab:deceptive_specific_prompt_replace_object}
\end{table}

\begin{table}[htbp]
\scriptsize
\centering
\begin{tabular}{@{}p{\linewidth}@{}}
\toprule
\textbf{Deceptive-Specific Prompt (\texttt{replace-attribute})}\\
\midrule
You will be given a caption describing the \{contents\_modality\}. Your task is to generate a hard negative caption based on the "attribute replacement" scenario using the criteria below:

~

***

[Generation Criteria]

1. Replace an adjective word in the given caption with a new adjective word that is not in the given caption.

2. Ensure the new caption has higher similarity to the \{contents\_modality\} in \{contents\_modality\}-text crossmodal model than the given caption.

3. Introduce a contradiction compared to the given caption, but avoid simple negations (e.g., using words like "no", "not", "empty", or "without").

4. Make fewer than \{max\_word\_distance\_plus\_one\} word-level changes (add, delete, or substitute words) to the given caption without fully rewriting it to generate the new caption.

~

[Given Caption]

- \{caption\}

***

~

Write only the new caption starting with "Generated Caption: ", without explanation. \\
\bottomrule
\end{tabular}
    \caption{Deceptive-specific prompt (\texttt{replace-attribute}).}
    \label{tab:deceptive_specific_prompt_replace_attribute}
\end{table}

\begin{table}[htbp]
\scriptsize
\centering
\begin{tabular}{@{}p{\linewidth}@{}}
\toprule
\textbf{Deceptive-Specific Prompt (\texttt{replace-relation})}\\
\midrule
You will be given a caption describing the \{contents\_modality\}. Your task is to generate a hard negative caption based on the "relation replacement" scenario using the criteria below:

~

***

[Generation Criteria]

1. Replace an action or a spatial relationship in the given caption with a new action or spatial relationship that is not in the given caption.

2. Ensure the new caption has higher similarity to the \{contents\_modality\} in \{contents\_modality\}-text crossmodal model than the given caption.

3. Introduce a contradiction compared to the given caption, but avoid simple negations (e.g., using words like "no", "not", "empty", or "without").

4. Make fewer than \{max\_word\_distance\_plus\_one\} word-level changes (add, delete, or substitute words) to the given caption without fully rewriting it to generate the new caption.

~

[Given Caption]

- \{caption\}

***

~

Write only the new caption starting with "Generated Caption: ", without explanation. \\
\bottomrule
\end{tabular}
    \caption{Deceptive-specific prompt (\texttt{replace-relation}).}
    \label{tab:deceptive_specific_prompt_replace_relation}
\end{table}

\begin{table}[htbp]
\scriptsize
\centering
\begin{tabular}{@{}p{\linewidth}@{}}
\toprule
\textbf{Deceptive-Specific Prompt (\texttt{replace-count})}\\
\midrule
You will be given a caption describing the \{contents\_modality\}. Your task is to generate a hard negative caption based on the "counting replacement" scenario using the criteria below:

~

***

[Generation Criteria]

1. Replace the numerical count of a key object in the given caption (e.g., from "two" to "three").

2. Ensure the new caption has higher similarity to the \{contents\_modality\} in \{contents\_modality\}-text crossmodal model than the given caption.

3. Introduce a contradiction compared to the given caption, but avoid simple negations (e.g., using words like "no", "not", "empty", or "without").

4. Make fewer than \{max\_word\_distance\_plus\_one\} word-level changes (add, delete, or substitute words) to the given caption without fully rewriting it to generate the new caption.

~

[Given Caption]

- \{caption\}

***

~

Write only the new caption starting with "Generated Caption: ", without explanation. \\
\bottomrule
\end{tabular}
    \caption{Deceptive-specific prompt (\texttt{replace-count}).}
    \label{tab:deceptive_specific_prompt_replace_count}
\end{table}

\begin{table}[htbp]
\scriptsize
\centering
\begin{tabular}{@{}p{\linewidth}@{}}
\toprule
\textbf{Deceptive-Specific Prompt (\texttt{add-object})}\\
\midrule
You will be given a caption describing the \{contents\_modality\}. Your task is to generate a hard negative caption based on the "object addition" scenario using the criteria below:

~

***

[Generation Criteria]

1. Generate a new plausible but uncommon object that's not in the given caption, and then add the new object to make a new caption.

2. Ensure the new caption has higher similarity to the \{contents\_modality\} in \{contents\_modality\}-text crossmodal model than the given caption.

3. Introduce a contradiction compared to the given caption, but avoid simple negations (e.g., using words like "no", "not", "empty", or "without").

4. Make fewer than \{max\_word\_distance\_plus\_one\} word-level changes (add, delete, or substitute words) to the given caption without fully rewriting it to generate the new caption.

~

[Given Caption]

- \{caption\}

***

~

Write only the new caption starting with "Generated Caption: ", without explanation. \\
\bottomrule
\end{tabular}
    \caption{Deceptive-specific prompt (\texttt{add-object}).}
    \label{tab:deceptive_specific_prompt_add_object}
\end{table}

\begin{table}[htbp]
\scriptsize
\centering
\begin{tabular}{@{}p{\linewidth}@{}}
\toprule
\textbf{Deceptive-Specific Prompt (\texttt{add-attribute})}\\
\midrule
You will be given a caption describing the \{contents\_modality\}. Your task is to generate a hard negative caption based on the "attribute addition" scenario using the criteria below:

~

***

[Generation Criteria]

1. Add a new plausible but uncommon attribute for the object in the given caption.

2. Ensure the new caption has higher similarity to the \{contents\_modality\} in \{contents\_modality\}-text crossmodal model than the given caption.

3. Introduce a contradiction compared to the given caption, but avoid simple negations (e.g., using words like "no", "not", "empty", or "without").

4. Make fewer than \{max\_word\_distance\_plus\_one\} word-level changes (add, delete, or substitute words) to the given caption without fully rewriting it to generate the new caption.

~

[Given Caption]

- \{caption\}

***

~

Write only the new caption starting with "Generated Caption: ", without explanation. \\
\bottomrule
\end{tabular}
    \caption{Deceptive-specific prompt (\texttt{add-attribute}).}
    \label{tab:deceptive_specific_prompt_add_attribute}
\end{table}

\begin{table}[htbp]
\scriptsize
\centering
\begin{tabular}{@{}p{\linewidth}@{}}
\toprule
\textbf{Deceptive-Specific Prompt (\texttt{swap-object})}\\
\midrule
You will be given a caption describing the \{contents\_modality\}. Your task is to generate a hard negative caption based on the "object swapping" scenario using the criteria below:

~

***

[Generation Criteria]

1. First locate two swappable nouns in the given caption, and then swap them to make a new caption (e.g., from "woman looking at elephant" to "elephant looking at woman")

2. Ensure the new caption has higher similarity to the \{contents\_modality\} in \{contents\_modality\}-text crossmodal model than the given caption.

3. Introduce a contradiction compared to the given caption, but avoid simple negations (e.g., using words like "no", "not", "empty", or "without").

4. Make fewer than \{max\_word\_distance\_plus\_one\} word-level changes (add, delete, or substitute words) to the given caption without fully rewriting it to generate the new caption.

~

[Given Caption]

- \{caption\}

***

~

Write only the new caption starting with "Generated Caption: ", without explanation. \\
\bottomrule
\end{tabular}
    \caption{Deceptive-specific prompt (\texttt{swap-object}).}
    \label{tab:deceptive_specific_prompt_swap_object}
\end{table}

\begin{table}[htbp]
\scriptsize
\centering
\begin{tabular}{@{}p{\linewidth}@{}}
\toprule
\textbf{Deceptive-Specific Prompt (\texttt{swap-attribute})}\\
\midrule
You will be given a caption describing the \{contents\_modality\}. Your task is to generate a hard negative caption based on the "attribute swapping" scenario using the criteria below:

~

***

[Generation Criteria]

1. First locate two swappable adjectives in the given caption describing different objects, and then swap them to make a new caption (e.g., from "a red apple and a purple grape" to "a purple apple and a red grape").

2. Ensure the new caption has higher similarity to the \{contents\_modality\} in \{contents\_modality\}-text crossmodal model than the given caption.

3. Introduce a contradiction compared to the given caption, but avoid simple negations (e.g., using words like "no", "not", "empty", or "without").

4. Make fewer than \{max\_word\_distance\_plus\_one\} word-level changes (add, delete, or substitute words) to the given caption without fully rewriting it to generate the new caption.

~

[Given Caption]

- \{caption\}

***

~

Write only the new caption starting with "Generated Caption: ", without explanation. \\
\bottomrule
\end{tabular}
    \caption{Deceptive-specific prompt (\texttt{swap-attribute}).}
    \label{tab:deceptive_specific_prompt_swap_attribute}
\end{table}

\subsection{Implementation Details}  
For generating new captions with LLMs, we apply nucleus sampling~\cite{holtzman2020nucleussampling} with $p=0.95$ and a temperature of $\tau=0.7$ across all LLMs, except for GPT-4o, where we use the default hyperparameters provided by the OpenAI API.  
For self-training LLMs, we use a batch size of 16, a LoRA~\cite{hu2022lora} rank of 16, a LoRA alpha of 32, and a learning rate of $2 \times 10^{-4}$.
Each LLM is trained for 3 epochs per round.
During multi-round training, we reset the LLM to its original checkpoint at the start of each round, rather than continuing from the last checkpoint, to mitigate overfitting~\cite{zelikman2022star,singh2024beyond}.
All experiments are conducted on a single NVIDIA RTX A6000 GPU.
All reported results are based on a single run per experiment.

\subsection{Human Evaluation}
\label{subsec:human_eval}

\begin{figure}[t]
    \centering
    \includegraphics[width=0.49\textwidth]{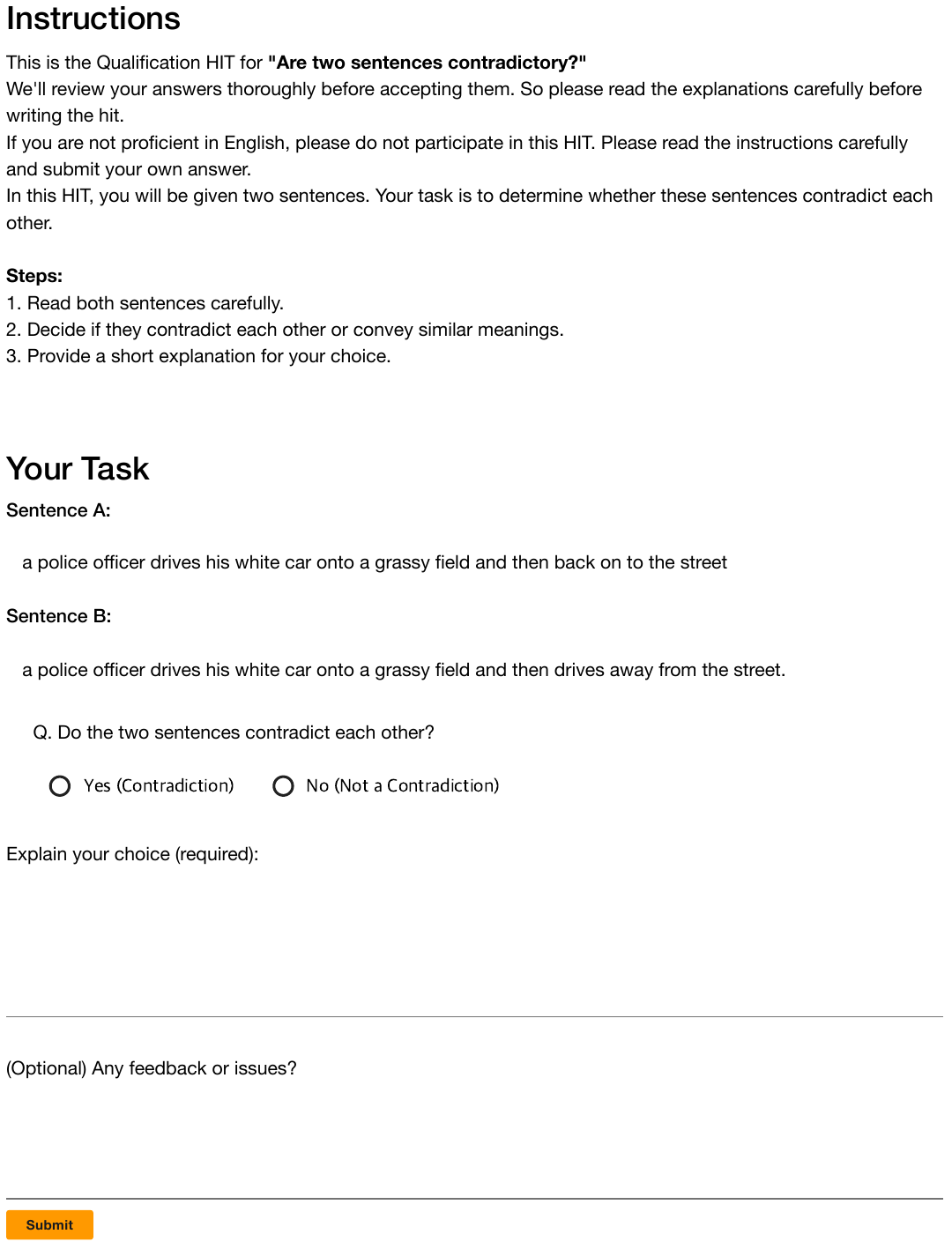}
    \caption{User interface for human evaluation: Task 1 (without video).}
    \label{fig:human_eval_task1}
\end{figure}

\begin{figure}[t]
    \centering
    \begin{subfigure}[b]{0.49\textwidth}
        \centering
        \includegraphics[width=\textwidth]{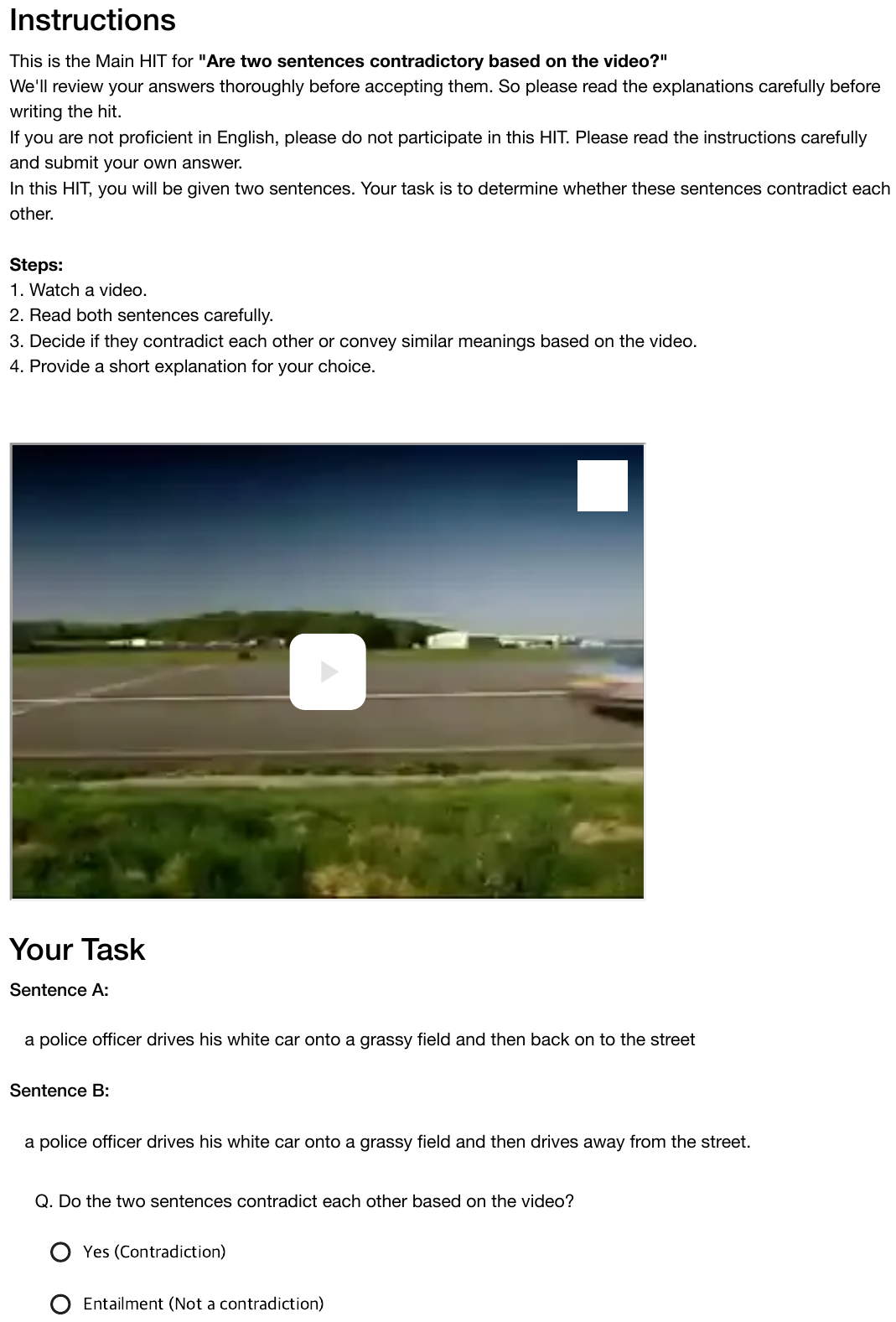}
    \end{subfigure}
    \hfill
    \begin{subfigure}[b]{0.49\textwidth}
        \centering
        \includegraphics[width=\textwidth]{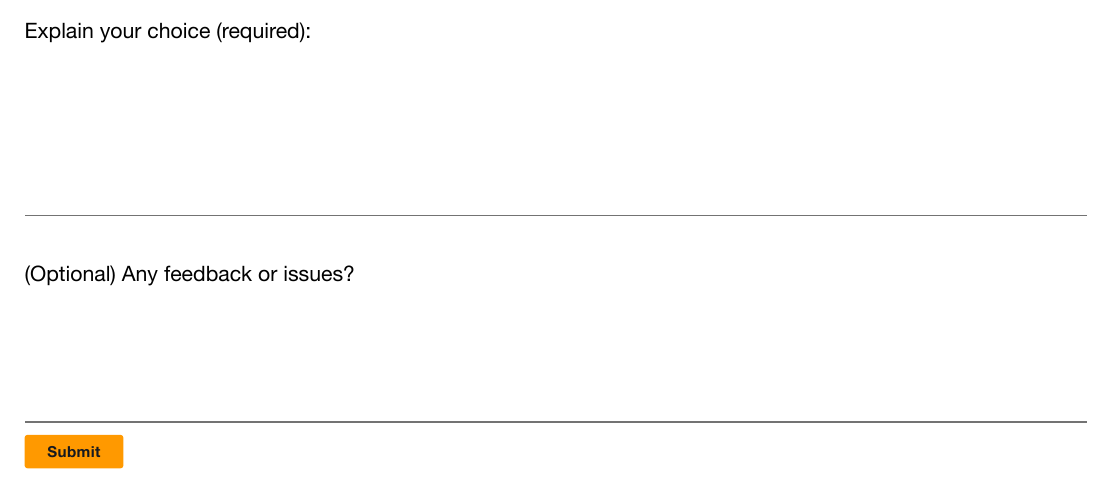}
    \end{subfigure}
    \caption{User interface for human evaluation: Task 2 (with video).}
    \label{fig:human_eval_task2}
\end{figure}

We provide a detailed explanation of the human evaluation process described in Sec.~\ref{subsec:performance_analysis} (Fig.~\ref{fig:analysis}-(c)).  
Two user interfaces were designed for evaluation on Amazon Mechanical Turk (AMT): one without video input (Fig.~\ref{fig:human_eval_task1}) and one with video input from MSRVTT (Fig.~\ref{fig:human_eval_task2}).  
For each data point, we collected five annotations to ensure reliability.
To maintain annotation quality, annotators were required to provide a short explanation for their responses.  
Additionally, we ensured that AMT workers were fairly compensated at approximately \$18 per hour (\$0.5 per HIT).

\section{Further Analyses}

\subsection{MAC Performance Across LLMs}
\label{subsec:mac_across_llms}

\begin{table}[t!] \begin{center}
    \small
   \begin{tabular}{lcccc}
    \toprule
    \multirow{2}{*}{\makecell[l]{Method}} & 
    \multicolumn{2}{c}{ASR$_\uparrow$} & 
    \multicolumn{2}{c}{Diversity$_\uparrow$} \\
    \cmidrule(r{0.3em}){2-3} \cmidrule(r{0.3em}){4-5}
    & \makecell{Cross} & \makecell{Total} & \makecell{$H$} & \makecell{$D_1$} \\
    \midrule
    Qwen-2.5-7B & 18.80 & 4.50 & 6.454 & 0.538 \\
    Llama-3.1-8B & \textbf{32.20} & 7.70 & \textbf{6.809} & \textbf{0.638} \\
    Gemma-2-9B & 19.80 & 8.30 & 6.472 & 0.507 \\
    Llama-3.1-70B & 20.80 & 9.10 & 6.416 & 0.520 \\
    GPT-4o$_\texttt{2024-08-06}$ & 21.10 & \textbf{14.40} & 6.440 & 0.502 \\
    \bottomrule
   \end{tabular}
   \caption{Attacking LanguageBind in MSRVTT test set with diverse LLMs ($N$=1). All LLMs use the deceptive-general prompt.}
   \label{tab:analysis_llm_selection}
\end{center}
\end{table}

We examine the applicability across different language models, such as Qwen 2.5~\cite{yang2024qwen25} and Gemma 2~\cite{team2024gemma2}, as well as 
GPT-4o~\cite{achiam2023gpt4}.
As shown in Table~\ref{tab:analysis_llm_selection}, larger or proprietary models do not necessarily lead to more effective deception.  
For instance, while GPT-4o achieves the highest ASR, its diversity is lower than that of Llama-3.1-8B. Moreover, Llama-3.1-8B with $N=4$ achieves a significantly higher ASR (24.80 in Table~\ref{tab:main_result_table}) compared to GPT-4o (14.40). This suggests that using a smaller model with a Best-of-$N(>1)$ approach is more effective than relying on a proprietary model with a budget of $N=1$.

\subsection{MAC Performance Across Generation Strategies}
\label{subsec:mac_across_ttc}

\begin{table}[t!] \begin{center}
    \small
   \begin{tabular}{llcccc}
    \toprule
    \multirow{2}{*}{\makecell[l]{Method}} &  &
    \multicolumn{2}{c}{ASR$_\uparrow$} & 
    \multicolumn{2}{c}{Diversity$_\uparrow$} \\
    \cmidrule(r{0.3em}){3-4} \cmidrule(r{0.3em}){5-6}
    & \makecell{Time} & \makecell{Cross} & \makecell{Total} & \makecell{$H$} & \makecell{$D_1$} \\
    \midrule
    \rowcolor{gray!15} \multicolumn{6}{l}{$N=4$}     \\
    Sequential & $O(N)$ & 38.50 & 20.10 & \textbf{6.809} & 0.658 \\
    Parallel & $O(1) $ & 42.40 & 24.80 & 6.808 & 0.626 \\
    \rowcolor{gray!15} \multicolumn{6}{l}{$N=8$}     \\
    Sequential & $O(N)$ & 45.40 & 28.50 & 6.764 & \textbf{0.675} \\
    Parallel & $O(1)$ & \textbf{49.20} & \textbf{36.40} & 6.773 & 0.601 \\
    \bottomrule
   \end{tabular}
   \caption{Attacking LanguageBind in MSRVTT test set with parallel/sequential generation in TTC with Best-of-$N$ budget. All methods use Llama-3.1-8B with the deceptive-general prompt.}
   \label{tab:analysis_TTC_selection}
\end{center}
\end{table}

LLMs can generate $N$ multiple candidates using two main approaches: sequential generation and parallel generation.  
Sequential generation involves iteratively refining responses based on the output from the previous turn~\cite{shinn2023reflexion,madaan2023selfrefine}, whereas parallel generation produces $N$ responses simultaneously without a refinement process.
While the sequential approach achieves slightly higher diversity in Table~\ref{tab:analysis_TTC_selection}, it underperforms parallel generation in terms of ASR.  
Additionally, sequential generation has a time complexity of $O(N)$, whereas parallel generation operates with a constant time complexity of $O(1)$.
This makes sequential generation less practical for self-training and inference, as it significantly increases computational overhead.  
Therefore, we adopt parallel generation as the default method for generating $N$ multiple candidates.

\begin{figure*}[t]
    \centering
    \includegraphics[width=\textwidth]{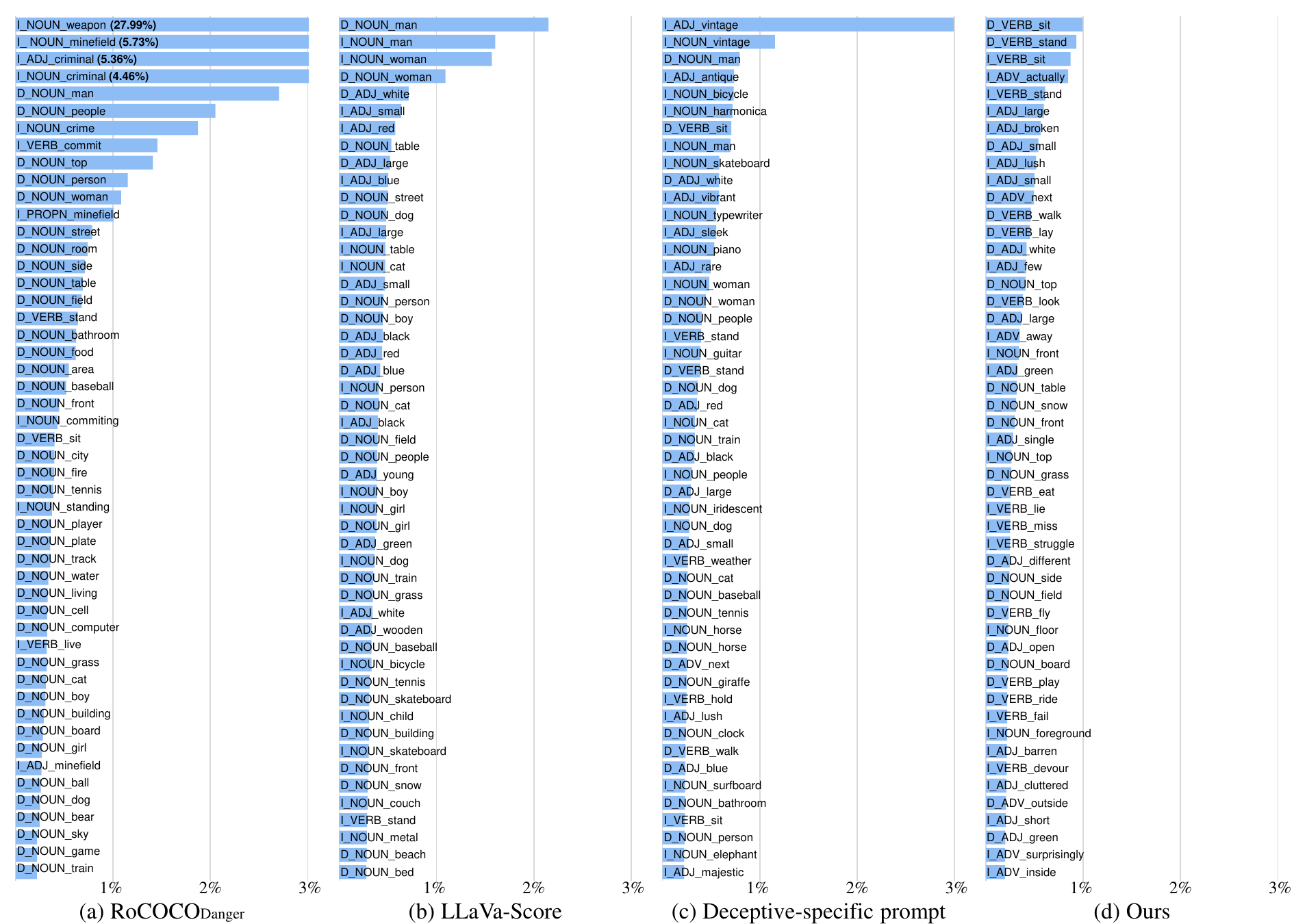}
    \caption{
Distribution of attribute-enhanced tokens from different methods.
}
    \label{fig:app_tokendist}
\end{figure*}

\subsection{Group-wise Diversity Analysis}  
\label{subsec:diversity_analysis}

Fig.~\ref{fig:app_tokendist} presents the distributions of attribute-enhanced tokens generated by different methods, including RoCOCO$_\text{Danger}$, LLaVa-Score, deceptive-specific prompt (zero-shot), and our diversity-promoted self-trained approach.  
Notably, in the first three methods, certain tokens appear with extremely high frequency. For instance, $\texttt{I\_NOUN\_weapon}$ occurs in more than 25\% of the generated outputs, while other frequent tokens like $\texttt{I\_ADJ\_vintage}$ exceed 3\%.
In contrast, our approach produces a much more balanced token distribution, with the most frequent token appearing in less than 1\% of cases.

\begin{table*}[t!] \begin{center}
    \begin{adjustbox}{width=\linewidth}
   \begin{tabular}{lcccccccc}
    \toprule
    \multirow{2}{*}{\makecell[l]{Method}} & 
    \multicolumn{5}{c}{ASR$_\uparrow$} & 
    \multicolumn{3}{c}{Diversity$_\uparrow$} \\
    \cmidrule(r{0.3em}){2-6} \cmidrule(r{0.3em}){7-9}
    & \makecell{Cross} & \makecell{Uni} & \makecell{Dist} & \makecell{Aux} & \makecell{Total} & \makecell{$H$} & \makecell{$\hat{H}$} & \makecell{$D_1$} \\
    \midrule
    \rowcolor{gray!15} \multicolumn{9}{l}{\textbf{N=1}}     \\
    Deceptive-General Prompt (zero-shot) & 32.20 & 40.80 & 74.90 & 98.10 & 7.70 & 6.809 & \underline{0.958} & \underline{0.638} \\
    \midrule
    \rowcolor{gray!15} \multicolumn{9}{l}{\textbf{N=4}}     \\
    Deceptive-General Prompt (zero-shot) & 42.40 & 56.50 & 80.90 & 97.90 & 24.80 & 6.808 & 0.953 & 0.626 \\
    + Self-Train & 48.90 & 75.80 & \underline{95.30} & \underline{99.90} & 39.70 & \underline{6.900} & 0.952 & 0.587 \\
    + Self-Train + Diversity-Promoted & 49.00 & \underline{77.00} & 94.00 & 99.80 & 40.60 & 6.882 & 0.953 & 0.598 \\
    + Self-Train + Large-$N$ Distilled & \underline{52.90} & \textbf{80.10} & 93.30 & \textbf{100.00} & \underline{44.20} & 6.839 & 0.951 & 0.594 \\
    + Self-Train + Large-$N$ Distilled + Diversity-Promoted (\textbf{Ours})& \textbf{53.50} & 76.60 & \textbf{95.50} & \textbf{100.00} & \textbf{45.60} & \textbf{7.125} & \textbf{0.965} & \textbf{0.667} \\
    \bottomrule
   \end{tabular}
   \end{adjustbox}
   \caption{Ablation study: Fine-grained attack evaluation on the MSRVTT test set for LanguageBind. The Self-Train method is applied with a single iteration.}
   \label{tab:ablation_study_msrvtt_lb}
\end{center}
\end{table*}

\subsection{Ablation Study}
\label{subsec:ablation_study}
We conduct an ablation study on our method using fine-grained metrics, as shown in Table~\ref{tab:ablation_study_msrvtt_lb}.  

\textbf{ASR.}  
As expected, setting $N=4$ improves cross-modal ASR by 10\% points and unimodal ASR by 15.7\% points, compared to $N=1$.  
Naïve self-training particularly enhances unimodal ASR (+19.3 \% points) and the distance-based criterion (+14.4 \% points), followed by cross-modal ASR (+6.5 \% points).  
Finally, self-training with large-$N$ and our final method further boost cross-modal ASR, achieving the highest total ASR.

\textbf{Diversity.}  
While standard self-training and large-$N$ self-training produce mixed results compared to the deceptive-general prompt (\eg higher entropy $H$ but lower normalized entropy $\hat{H}$ and distinct-1 $D_1$), our diversity-promoting self-training with large-$N$ consistently outperforms the deceptive-general prompt across all diversity metrics.

\subsection{Multi-round Self-training}
\label{subsec:multi_round_self_train}

\begin{figure}[t]
    \centering
    \includegraphics[width=0.49\textwidth]{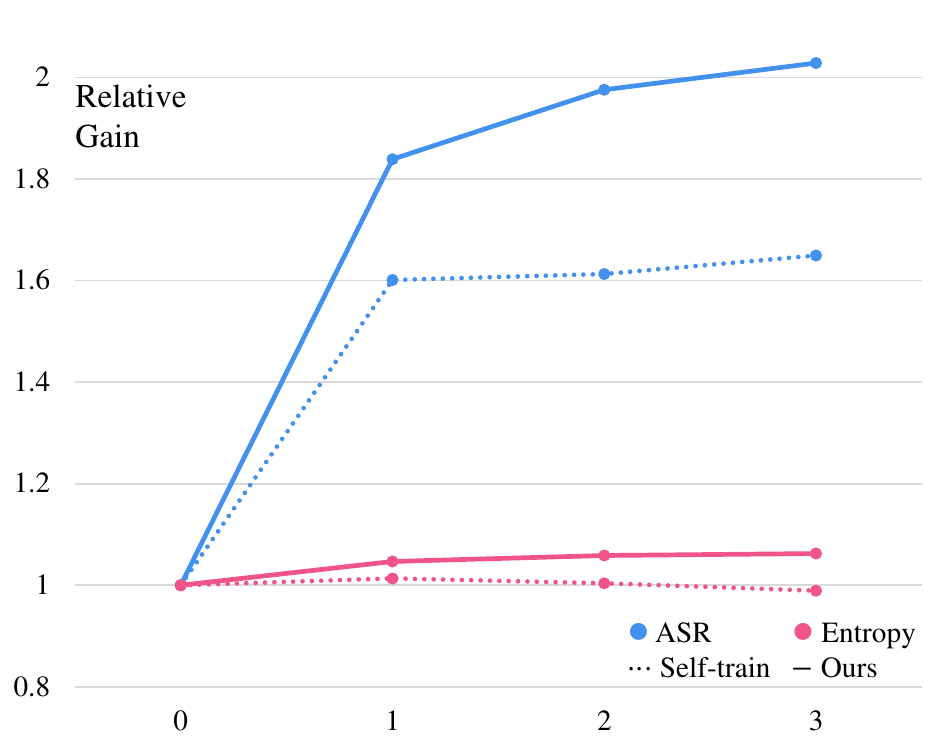}
    \caption{Influence of multi-round self-training in MSRVTT.}
    \label{fig:analysis_msrvtt}
\end{figure}

In addition to the results on AudioCaps shown in Fig.~\ref{fig:analysis}-(b), we further evaluate multi-round self-training on MSRVTT, as demonstrated in Fig.~\ref{fig:analysis_msrvtt}.
Similarly, the results demonstrate that our approach achieves a significant improvement in ASR, yielding over a 2× relative gain by the third round.  
Moreover, while entropy typically decreases with self-training, our approach continues to show consistent improvement, indicating sustained diversity enhancement across different datasets.

\begin{table*}[t!]
\centering
\begin{adjustbox}{width=\linewidth}
\begin{tabular}{lcccccccccccc}
\toprule
\multirow{3}{*}{\makecell[l]{Method}}
& \multicolumn{4}{c}{\textbf{(a) Image (CLIP/Flickr30K)}} & \multicolumn{4}{c}{\textbf{(b) Video (LB/LSMDC)}} & \multicolumn{4}{c}{\textbf{(c) Audio (LB/Clotho)}} \\
& \multicolumn{2}{c}{ASR$_\uparrow$} & \multicolumn{2}{c}{Diversity$_\uparrow$} & \multicolumn{2}{c}{ASR$_\uparrow$} & \multicolumn{2}{c}{Diversity$_\uparrow$} & \multicolumn{2}{c}{ASR$_\uparrow$} & \multicolumn{2}{c}{Diversity$_\uparrow$} \\
\cmidrule(r{0.3em}){2-3} \cmidrule(r{0.3em}){4-5} \cmidrule(r{0.3em}){6-7} \cmidrule(r{0.3em}){8-9} \cmidrule(r{0.3em}){10-11} \cmidrule(r{0.3em}){12-13}
& Cross & Total & $H$ & $D_1$
& Cross & Total & $H$ & $D_1$
& Cross & Total & $H$ & $D_1$ \\
\midrule
\rowcolor{gray!15}
\multicolumn{13}{l}{\textbf{N=1}} \\
Deceptive-General Prompt (zero-shot) & 23.70 & 6.12 & 7.437 & \underline{0.290} & 39.90 & 15.20 & 6.842 & \underline{0.642} & 34.97 & 14.18 & 7.158 & \underline{0.225} \\
\midrule
\rowcolor{gray!15}
\multicolumn{13}{l}{\textbf{N=4}} \\
Deceptive-General Prompt (zero-shot) & 32.90 & 17.42 & 7.479 & \underline{0.290} & 54.70 & 37.30 & \underline{6.922} & 0.632 & 50.37 & 36.15 & \underline{7.174} & 0.217 \\
+ Self-Train & 39.04 & 29.34 & 7.350 & 0.285 & 58.30 & 50.70 & 6.788 & 0.585 & 54.07 & 44.08 & 7.017 & 0.201 \\
+ Self-Train + Large-$N$ Distilled & \textbf{41.88} & \underline{33.66} & \underline{7.489} & 0.287 & \textbf{61.40} & \underline{54.20} & 6.841 & 0.575 & \underline{57.51} & \underline{47.90} & 7.061 & 0.200 \\
+ Self-Train + Large-$N$ Distilled + Diversity-Promoted (\textbf{Ours}) & \underline{41.82} & \textbf{34.42} & \textbf{7.716} & \textbf{0.314} & \underline{61.30} & \textbf{54.80} & \textbf{7.141} & \textbf{0.655} & \textbf{57.72} & \textbf{49.09} & \textbf{7.410} & \textbf{0.233} \\
\bottomrule
\end{tabular}
\end{adjustbox}
\caption{
Additional results on diverse datasets using Llama-3.1-8B: Flickr30K, LSMDC, Clotho.
}
\label{tab:results_on_diverse_datasets}
\end{table*}

\subsection{MAC Performance Across Diverse Configurations}

\begin{table*}[t!]
\centering
\begin{adjustbox}{width=\linewidth}
\begin{tabular}{lcccccccc}
\toprule
\multirow{3}{*}{\makecell[l]{Method}}
& \multicolumn{4}{c}{\textbf{Audio (LB/AudioCaps)}} & \multicolumn{4}{c}{\textbf{Audio (CLAP/AudioCaps)}} \\
& \multicolumn{2}{c}{ASR$_\uparrow$} & \multicolumn{2}{c}{Diversity$_\uparrow$} & \multicolumn{2}{c}{ASR$_\uparrow$} & \multicolumn{2}{c}{Diversity$_\uparrow$} \\
\cmidrule(r{0.3em}){2-3} \cmidrule(r{0.3em}){4-5} \cmidrule(r{0.3em}){6-7} \cmidrule(r{0.3em}){8-9}
& Cross & Total & $H$ & $D_1$
& Cross & Total & $H$ & $D_1$ \\
\midrule
\rowcolor{gray!15}
\multicolumn{9}{l}{\textbf{N=4}} \\
Deceptive-General Prompt (zero-shot) & 42.60 & 29.02 & 6.566 & 0.172 & 37.65 & 24.07 & \textbf{6.852} & \underline{0.173} \\
+ Self-Train & 55.37 & 47.35 & 6.472 & 0.157 & 36.45 & 29.98 & 6.478 & 0.160 \\
+ Self-Train + Large-$N$ Distilled & \underline{58.38} & \underline{51.57} & 6.508 & 0.157 & \underline{38.33} & \underline{32.70} & 6.476 & 0.159 \\
+ Self-Train + Large-$N$ Distilled + Diversity-Promoted (\textbf{Ours}) & \textbf{60.25} & \textbf{52.87} & \textbf{6.868} & \textbf{0.191} & \textbf{38.41} & \textbf{33.11} & \underline{6.829} & \textbf{0.186} \\
\bottomrule
\end{tabular}
\end{adjustbox}
\caption{
Attacking LanguageBind/CLAP in AudioCaps test set using Llama-3.1-8B.
}
\label{tab:lb_vs_clap}
\end{table*}

Beyond the COCO, MSRVTT, and AudioCaps datasets, we further explore other datasets: Flickr30K~\cite{young2014flickr30k} for image-text, LSMDC~\cite{rohrbach2017lsmdc} for video-text, and Clotho~\cite{drossos2020clotho} for audio-text.

For Flickr30K, we adopt the Karpathy test split~\cite{karpathy2017imagecap} as the test set, which consists of 1,000 images paired with 5,000 captions. The train set contains 29,000 images and 145,000 captions.
For LSMDC, we utilize the test split from~\citet{li2023unmasked}, which includes 1,000 videos, each associated with a single caption. The train set contains 101,020 videos with 101,020 captions.
For Clotho, we use the test split from~\citet{oncescu2021audioretrieval}, which consists of 1,045 audio clips with 5,225 captions. The train set includes 2,314 audios with 11,570 captions.

Table~\ref{tab:results_on_diverse_datasets} shows that LLMs effectively deceive the target representations across diverse datasets.
Furthermore, our method consistently outperforms baseline methods in terms of both ASR and diversity.

Lastly, to demonstrate that MAC can be readily extended to other target models, we evaluate the performance of our framework using CLAP~\cite{wu2023clap} as the target model for the audio-text dataset and compare the results with LanguageBind.
As shown in Table~\ref{tab:lb_vs_clap}, we observe that the trends confirmed in the LanguageBind-based experiments are also evident in the CLAP-based experiments. However, CLAP exhibits consistently lower ASR across all metrics. We presume this occurs because LanguageBind, which binds multiple modalities at once, may expose greater vulnerability compared to models that focus exclusively on audio-text alignment.

\subsection{MAC Performance Across Long Captions}
\label{subsec:mac_on_long_captions}

\begin{table*}[t!]
\centering
\begin{adjustbox}{width=\linewidth}
\begin{tabular}{lcccccc}
\toprule
\multirow{3}{*}{\makecell[l]{Method}}
& \multicolumn{3}{c}{\textbf{Image (CLIP/ImageParagraph)}} & \multicolumn{3}{c}{\textbf{Video (LB/ActivityNet)}} \\
& \multicolumn{2}{c}{ASR$_\uparrow$} & Diversity$_\uparrow$ & \multicolumn{2}{c}{ASR$_\uparrow$} & Diversity$_\uparrow$ \\
\cmidrule(r{0.3em}){2-3} \cmidrule(r{0.3em}){4-4} \cmidrule(r{0.3em}){5-6} \cmidrule(r{0.3em}){7-7}
& Cross & Total & $H$
& Cross & Total & $H$ \\
\midrule
\rowcolor{gray!15}
\multicolumn{7}{l}{\textbf{N=4}} \\
Deceptive-General Prompt (zero-shot) & 26.56 & 4.82 & 6.651 & 40.23 & 6.07 & 7.306 \\
\midrule
\rowcolor{gray!15}
\multicolumn{7}{l}{\textbf{N=16}} \\
Deceptive-General Prompt (zero-shot) & 33.71 & 14.34 & 6.822 & 46.42 & 16.80 & 7.474 \\
+ Self-Train + Large-$N$ Distilled + Diversity-Promoted (\textbf{Ours}) & \textbf{57.98} & \textbf{48.45} & \textbf{6.983} & \textbf{67.10} & \textbf{54.78} & \textbf{7.777} \\
\bottomrule
\end{tabular}
\end{adjustbox}
\caption{Results on long captions: Stanford Image Paragraph and ActivityNet Captions. We used $N=32$ for the Large-$N$.}
\label{tab:results_long_captions}
\end{table*}

\begin{figure}[t]
    \centering
    \includegraphics[width=\columnwidth]{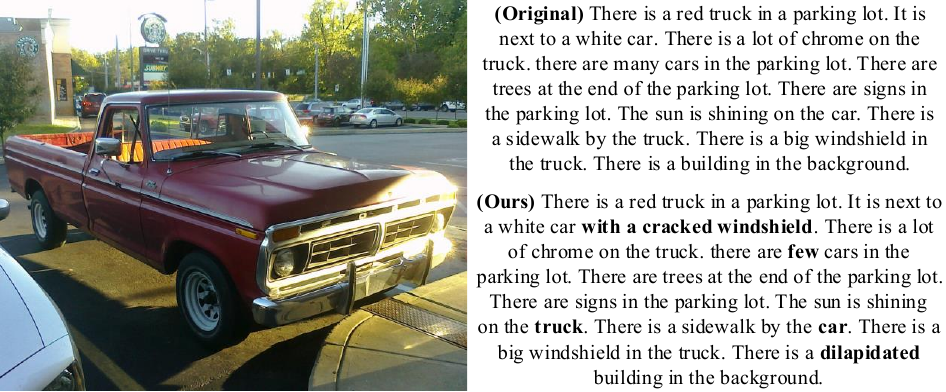}
    \caption{Qualitative examples for MAC on Stanford Image Paragraph test set. \textbf{Bold} phrases denote text updates.}
    \label{fig:long_caption}
\end{figure}

We further extend our benchmark with long captioning corpora by exploring two different data sources: Stanford Image Paragraph~\cite{krause2017hierarchical} for image-text and ActivityNet Captions~\cite{krishna2017dense} for video-text, whose average word lengths are 60 and 48, respectively.
Following \citet{Zhang2018ECCV, gabeur2020multi}, we aggregate all sentences from each video in chronological order to obtain long captions from ActivityNet captions.

For Stanford Image Paragraph, the test set consists of 2,489 images paired with 2,489 captions. The train set contains 14,575 images and 14,575 captions.
For ActivityNet Captions, the test split includes 4,429 videos, each associated with a single caption. The train set contains 9,032 videos with 9,032 captions.

Table~\ref{tab:results_long_captions} summarizes the results of long caption scenarios, where we can observe similar results with the short caption setup (\ie COCO and MSRVTT).

For a more comprehensive view of our benchmark for longer text inputs, we further share a qualitative example that successfully deceived CLIP from Stanford Image Paragraph in Fig.~\ref{fig:long_caption}.

\subsection{MAC Performance on Vision Language Models}


\begin{table}[t!]
\centering
\begin{adjustbox}{width=\linewidth}
\begin{tabular}{lccccc}
\toprule 
ASR$_\text{Total}$& \;\;CLIP\;\; & \;SigLIP\; & NegCLIP & \;\;BLIP\;\; & \;LLaVA\; \\
\midrule
\rowcolor{gray!15}
\multicolumn{6}{l}{\textbf{N=4}} \\
Zero-shot & 19.19 & 19.72 & 20.71 & 14.70 & 15.30 \\
\textbf{Ours} & \textbf{42.10} & \textbf{41.04} & \textbf{40.81} & \textbf{32.50} & \textbf{36.38} \\
\bottomrule
\end{tabular}
\end{adjustbox}
\caption{
Attacking five target models in COCO test set using Llama-3.1-8B.
}
\label{tab:mac_on_vlm}
\end{table}

In Table~\ref{tab:analysis_clip_transfer}, we show that LLMs such as Llama-3.1-8B can successfully deceive pre-trained multimodal representations, including CLIP, SigLIP, NegCLIP, and BLIP in COCO.
To further extend these pre-trained multimodal representations to recent vision language models (VLMs), we include LLaVA-1.5-7B\footnote{\href{https://huggingface.co/llava-hf/llava-1.5-7b-hf}{llava-hf/llava-1.5-7b-hf}}~\cite{liu2023llava,liu2024llava15} as a target representation.
Following \citet{li2024llavascore}, we adapt LLaVa-1.5-7B as an image-text matching score calculator by employing the following prompt format:  
\begin{quote}
    \textit{``Does this image $I$ match the following caption $T$? Answer Yes or No directly.''}
\end{quote}
Then, we extract the logits associated with the responses ``Yes'' and ``No'' for the next word prediction. We then define the matching score as:
\begin{equation}
\text{score} = \frac{e^{P(\text{Yes} \mid \text{prompt})}}{e^{P(\text{Yes} \mid \text{prompt})} + e^{P(\text{No} \mid \text{prompt})}}
\end{equation}

As shown in Table~\ref{tab:mac_on_vlm}, LLaVA-1.5-7B surprisingly demonstrates a high susceptibility to deception, performing even worse than ``smaller'' BLIP in our experiments on COCO (ASR 36.38\% vs. 32.50\%).
Even without self-training, the ASR remains at 15.30\%, indicating that LLaVA-1.5-7B possesses inherent compositional vulnerabilities, too.
These findings suggest that recent VLMs can be deceived by carefully crafted text inputs, underscoring a critical challenge in their robustness.

\subsection{Qualitative Results}
\label{subsec:qualitative_results}





\begin{figure}[t]
    \centering
    \begin{subfigure}[b]{0.49\textwidth}
        \centering
        \includegraphics[width=\textwidth]{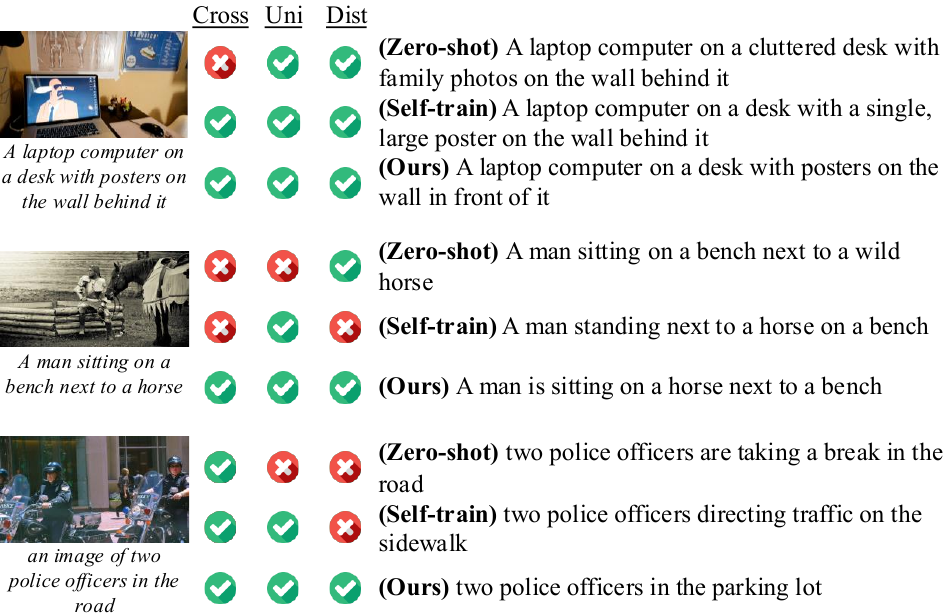}
        \caption{Qualitative examples on COCO.}
    \end{subfigure}
    \hfill
    \begin{subfigure}[b]{0.49\textwidth}
        \centering
        \includegraphics[width=\textwidth]{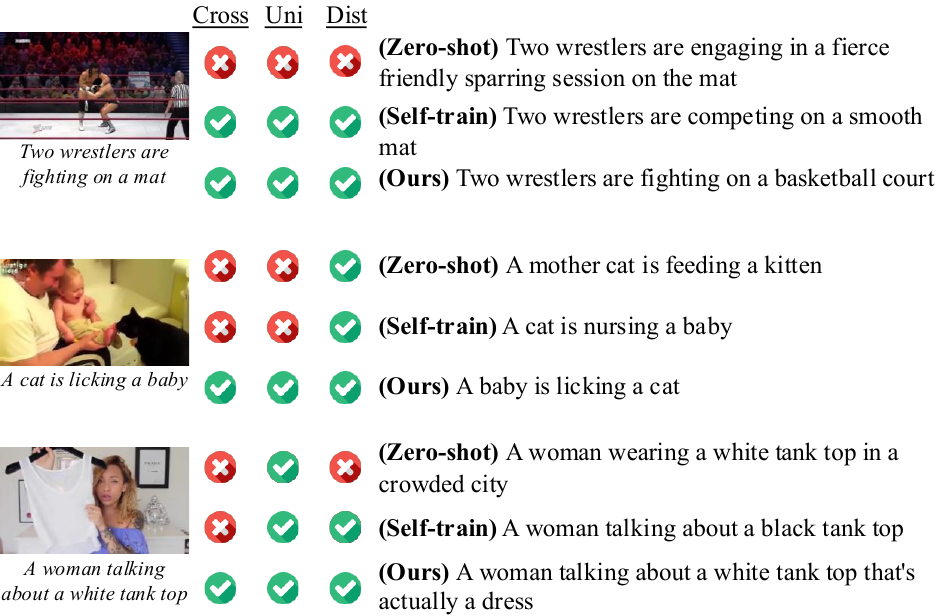}
        \caption{Qualitative examples on MSRVTT.}
    \end{subfigure}
    \hfill
    \begin{subfigure}[b]{0.49\textwidth}
        \centering
        \includegraphics[width=\textwidth]{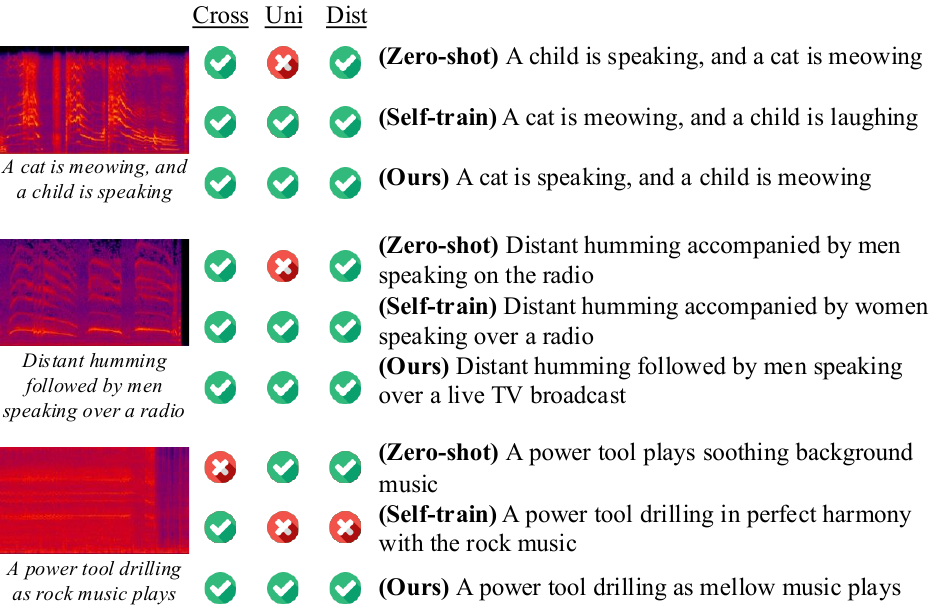}
        \caption{Qualitative examples on AudioCaps.}
    \end{subfigure}
    \hfill
    \begin{subfigure}[b]{0.49\textwidth}
        \centering
        \includegraphics[width=\textwidth]{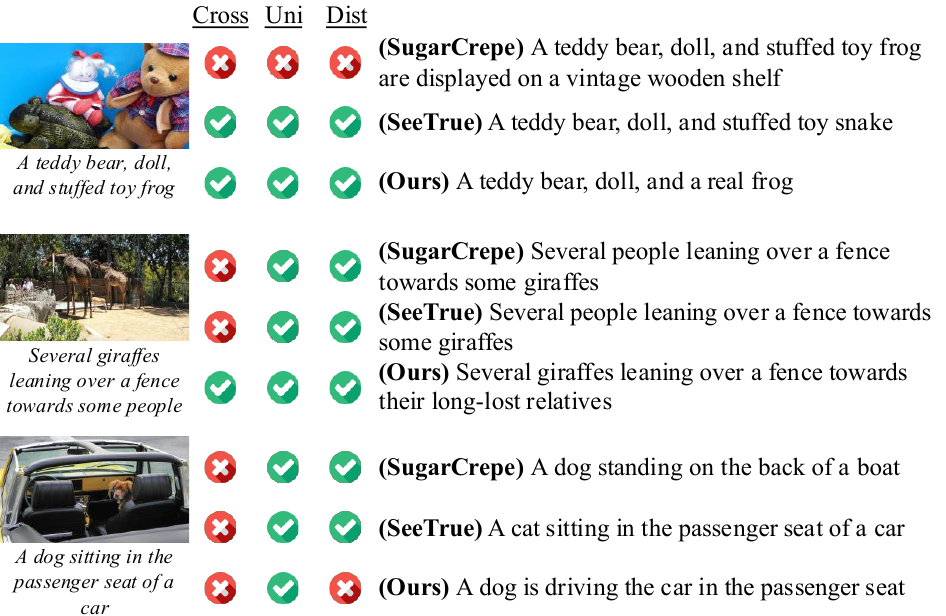}
        \caption{Comparison of prior approaches on COCO.}
    \end{subfigure}
    \caption{More qualitative examples.}
    \label{fig:app_qual}
\end{figure}

Fig.~\ref{fig:app_qual}-(a), Fig.~\ref{fig:app_qual}-(b), and Fig.~\ref{fig:app_qual}-(c) compare generated samples from different variants of our method across image, video, and audio modalities.
Additionally, Fig.~\ref{fig:app_qual}-(d) presents a comparison between our method and prior works (\ie SugarCrepe, SeeTrue).
Compared to other variants and prior arts, our self-training method effectively applies diverse modifications without being constrained to specific patterns.


\end{document}